\pdfoutput=1

\documentclass[11pt]{article}

\usepackage[]{acl}

\usepackage{times}
\usepackage{latexsym}

\usepackage[T1]{fontenc}

\usepackage[utf8]{inputenc}

\usepackage[scaled=0.8]{beramono}

\usepackage{microtype}
\usepackage{microtype}
\usepackage{url}
\usepackage[T1]{fontenc}
\usepackage{amsmath}
\usepackage{amssymb}
\usepackage{tabularx}
\usepackage{mathtools}
\usepackage{booktabs}
\usepackage{longtable}
\usepackage{tabu}
\usepackage{multirow}
\usepackage{amsfonts}
\usepackage{algorithm}
\usepackage{bbm}
\usepackage{subfigure}
\usepackage[noend]{algpseudocode}
\usepackage[normalem]{ulem}
\usepackage{enumitem}
\usepackage{tikz}
\usetikzlibrary{shapes.geometric}
\usetikzlibrary{patterns}
\usepackage{pgfplots}
\usetikzlibrary{backgrounds}
\usetikzlibrary{matrix,calc}
\def\BState{\State\hskip-\ALG@thistlm}
\usepackage{bm}
\usepackage{xcolor}

\usepackage{tikz-dependency}
\usetikzlibrary{patterns}

\usepackage[T2A,LGR,T1]{fontenc}
\usepackage[utf8]{inputenc}
\usepackage[russian,greek,spanish,es-nodecimaldot,english]{babel}
\usepackage[normalem]{ulem}
\title{\textsc{AutoLEX}: An Automatic Framework for Linguistic Exploration}

 \author{Aditi Chaudhary$^\dagger$, Zaid Sheikh$^\dagger$, David R Mortensen$^\dagger$, Antonios Anastasopoulos$^\ddagger$, Graham Neubig$^\dagger$ \\
   $^\dagger$Carnegie Mellon University, $^\ddagger$George Mason University \\
    \texttt{\{aschaudh,zsheikh,dmortens,gneubig\}@cs.cmu.edu} \hspace{.5cm} \texttt{antonis@gmu.edu}}

\begin{document}
\maketitle
\begin{abstract}
Each language has its own complex systems of word, phrase, and sentence construction, the guiding principles of which are often summarized in grammar descriptions for the consumption of linguists or language learners. 
However, manual creation of such descriptions is a fraught process, as creating descriptions which describe the language in ``its own terms'' without bias or error requires both a deep understanding of the language at hand and linguistics as a whole.
We propose an automatic framework \textsc{AutoLEX} that aims to ease linguists' discovery and extraction of concise descriptions of linguistic phenomena. 
Specifically, we apply this framework to extract descriptions for three  phenomena: \emph{morphological agreement, case marking,} and \emph{word order}, across several languages.
We evaluate the descriptions with the help of language experts and propose a method for automated evaluation when human evaluation is infeasible.\footnote{Code and data are released on \url{https://github.com/Aditi138/auto-lex-learn/tree/master/code}. 
Currently, the online web site (\url{https://aditi138.github.io/auto-lex-learn/index.html}) shows the rules for different languages.}
\end{abstract}

\section{Introduction}
\label{sec:intro}
Languages are amazingly diverse,  consisting of different systems for word formation (\emph{morphology}), phrase construction (\emph{syntax}), and meaning (\emph{semantics}).
These systems are governed by a set of guiding principles, referred to as \emph{grammar}.
Creating a human-readable description that highlights salient points of a language is one of the major endeavors undertaken by linguists. 
Such descriptions form an indispensable component of language documentation efforts, particularly for endangered or threatened languages \citep{himmelmann1998documentary,hale1992endangered,moseley2010atlas}. 
Furthermore, if descriptions can be created in a machine-readable format they can be used for developing language technologies \cite{pratapa-etal-2021-evaluating}.

Linguists and researchers have undertaken initiatives to collect linguistic properties  in a machine-readable format across several languages, 
WALS \cite{wals} being a standing example.
For instance, WALS can tell us that English objects occur after verbs, or that Turkish pronouns have symmetrical case. 
However, because WALS presents these properties across many diverse languages, these properties are necessarily defined at a coarse-grained level and cannot capture language-specific nuances.
WALS does not inform us of any exceptions to its general rules (e.g.~the cases when English objects come before verbs), and there are many aspects that are not even covered (e.g.~when a Turkish pronoun takes the accusative marker and when the nominative). 
There are other challenges to creating detailed descriptions, as for many of the 6,500+ languages, there are few or no formally trained linguists. 
Even in the ideal case where there is such a linguist, there are a plethora of linguistic phenomena to be covered, and it is hard to enumerate every single one through introspection.

\begin{figure*}
    \centering
    \includegraphics[width=\textwidth]{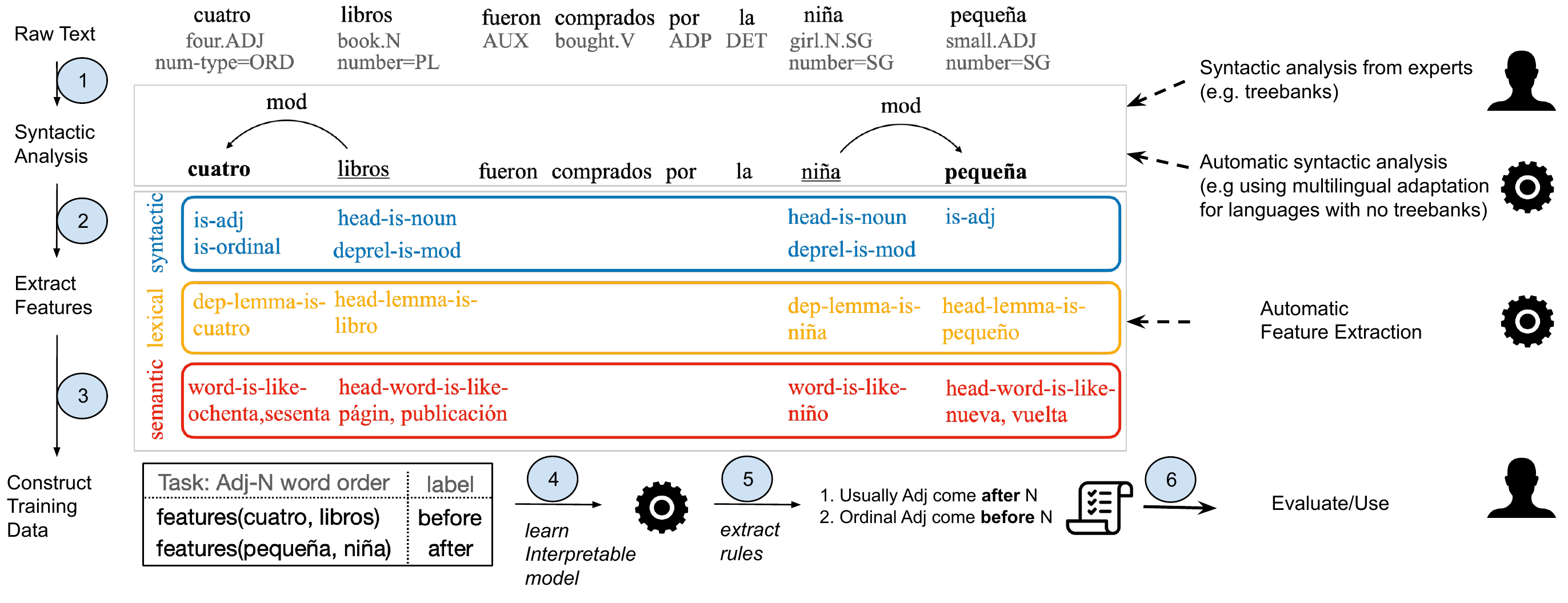}
    \caption{An overview of the \textsc{AutoLEX} framework  (with Adj-N order in Spanish as an example). The example sentence translates to \emph{Four books were bought by the small girl}. First, we formulate a linguistic question (e.g.~regarding Adj-N order) as a binary classification task (e.g.~``whether the Adj comes before/after the N'). Next, we perform syntactic analysis on the raw text, from which we extract syntactic, lexical, and semantic features to construct the training data. Finally, we learn an interpretable model from which we  extract concise rules. 
    }
    \label{fig:systemoverview}
    \vspace{-1em}
\end{figure*}

Thanks to the NLP advances, it is now possible to automate some \emph{local} aspects of linguistic analysis such as POS tagging \citep{toutanvoa-manning-2000-enriching}, dependency parsing \citep{kiperwasser-goldberg-2016-simple} or morphological analysis \citep{malaviya-etal-2018-neural}.
Recent advances in transfer learning have shown that this is possible to an extent, even for under-resourced languages \cite{kondratyuk-straka-2019-75}.
 A small amount of prior work has proposed  methods for answering specific questions about language, such as the analysis of word order \citep{ostling-2015-word,wang-eisner-2017-fine} and morphological agreement \citep{chaudhary-etal-2020-automatic}, or grammar extraction from inter-linear glosses \citep{bender-etal-2002-grammar} (\autoref{tab:relatedwork} in Appendix A compares the   questions answered by our and related work).

In this work, we propose \textsc{AutoLEX}, an automatic framework to aid linguistic exploration and description, with the goal of helping linguists develop fine-grained understanding of different linguistic phenomena.
The framework allows the linguist to ask a question such as ``what are the rules of object-verb order?'', or ``when do pronouns take the accusative case in Turkish?'', and automatically acquire first-pass answers.
\textsc{AutoLEX} analyses the texts in the corresponding languages and finds answers such as in English ``typical declarative constructions show VO, but interrogative sentences can show OV'', or in Turkish ``object pronouns take the accusative case.''
Specifically, we follow a multi-step process, as shown in  \autoref{fig:systemoverview}.
First, we define the linguistic question as a classification task (e.g.~``does the adjective come before the noun or not'';~\autoref{sec:questions}).
Second, we automatically extract syntactic, semantic, and surface-level features that may be predictive of the answer to this question (\autoref{sec:traindata}).
Next, we construct the training data and train an interpretable classifier such as a decision tree to identify the underlying patterns that answer this question. Finally, we extract and visualize interpretable rules (\autoref{sec:learning}).
This methodology is inspired by previous work on discovering fine-grained distinctions for individual phenomena \citep{wang-eisner-2017-fine,chaudhary-etal-2020-automatic}, 
but is significantly more general in that 
we demonstrate its ability to discover interesting features for word order, case marking, and morphological agreement.

We experiment with 61 languages for which we design an automated evaluation protocol which informs us how successful our framework is in discovering valid grammar rules~(\autoref{sec:goldexptauto}).
We further conduct a user study with linguists to evaluate how correct, readable, and novel the rules are perceived to be~(\autoref{sec:goldexptmanual}).
Finally, we apply this framework to a threatened language variety, Hmong Daw~(\texttt{mww}), and evaluate how well our framework extracts rules under zero-resource conditions~(\autoref{sec:lowexpt}).

\section{Formalizing Linguistic Questions}
\label{sec:questions}
The first step in applying \textsc{AutoLEX} to answer a  question is to determine whether we can formulate it as a classification task, with training data $\{ \langle \mathbf{x_1}, y_1 \rangle, \langle \mathbf{x_2},  y_2 \rangle \cdots, \langle \mathbf{x_n}, y_n \rangle\}$,
where $\mathbf{x_i}\!\in\!X$ are the input features and $y_i\!\in\!Y$ are the labels indicating the linguistic phenomenon.
Below, we describe how we define $Y$ for each phenomena, and discuss how to construct $X$ in the following section.
We use the UD schema \cite{mcdonald-etal-2013-universal} for representing the syntax and morphology. 

\paragraph{Case Marking} is a system of ``marking  syntactic dependents for the type of grammatical relation (subject, object, etc.) they bear to their heads'' \citep{blake2009history}.
Although there are different theories on how to formalize case marking,  we commit to the viewpoint that there are two types of cases: \emph{abstract} and \emph{morphological}, where 
the former is a universal property and the latter is its overt realization~
\citep{chomsky1993lectures, halle1993distributed}.
Thus, we formulate the explanation of case marking as determining \emph{when a word class  (e.g. nouns) marks a particular case  (e.g. nominative, etc.).}
Formally, for each POS tag $t$ we learn a separate model, where 
the input examples $x_i$ are the words having POS tag $t$ with the case feature marked (e.g. Case=Nominative). 
The model is trained to predict an output label ($y_i \in Y$), where $Y$ is the  label set of all observed case values  for that language.

\paragraph{Word Order} describes the relative position of the syntactic elements \citep{wo}, and is one of the major axes of linguistic description appearing in grammar sketches or databases such as WALS.
We consider the following five WALS relations $R$: subject-verb (82A),	 object-verb (83A), adjective-noun (87A),	 adposition-noun (85A)	and numeral-noun (89A).
In contrast to WALS, which only provides a single canonical order for the entire language, we pose the linguistic question as determining \emph{when does one word in such a relation appear before or after the other}.
Formally, the pair of words involved in the syntactic relation $\langle w_i^a, w_i^b \rangle\!\in\!r$ form the input example $x_i$ and the output label $y_i\!\in\!Y$ where $Y\!=\!\{\text{before}, \text{after}\}$.

\paragraph{Agreement} is the process where one word or morpheme selects a morphological form that agrees with that of another word/phrase in the sentence \cite{corbett2003agreement}.
We follow a similar problem formulation as \citet{chaudhary-etal-2020-automatic}, which asks the question \emph{when is agreement required between a head ($w_h$) and its dependent ($w_d$) for a morphological attribute $m$}.
We focus on the morphological attributes $M\!=\!\{\text{gender, person, number}\}$, which more often show agreement than other attributes \cite{corbett2009agreement}, 
and train a separate model for each. 
The pair of head-dependent words which both mark the morphological property $m$ form the input example $x_i$ and the output labels ($y_i$) are binary denoting if  agreement is observed or not between the  pair. 


\section{Feature Extraction}
\label{sec:traindata}

Now that we have provided three examples of converting linguistic questions into classification tasks, we design  features to help predict each question's answer.
We use linguistic knowledge to design features, but the feature extraction itself is automatic.
For a different question or language, a linguist can begin the process by using these initial features or even design new features as they deem fit. 
In step-2 of \autoref{fig:systemoverview}, we demonstrate example features extracted from a Spanish sentence for training the adjective-noun word order model.
We refer to the words participating in an input $x_i$ as \emph{focus words}.
These include the words describing the relation itself (e.g. the adjective \emph{cuatro} and its noun \emph{libros}) and also their respective heads and dependents.

\paragraph{Syntactic Features}
Prior work  \citep{blake2009history, kittila2011introduction, corbett2003agreement} has discussed the role of syntax and  morphology being important for determining the case and agreement.
In \autoref{fig:systemoverview}, we show a subset of features extracted for some of the focus words. For example, for the adjective, we derive  features from its POS tag (e.g.~``is-adj''), all of its morphological tags (e.g.~``is-ordinal'') and the dependency relation it is involved in (e.g.~``deprel-is-mod'').
We extract similar features for the adjective's head, which  is \emph{libros} (e.g.~``head-is-noun'').

\paragraph{Lexical Features}
An influential family of linguistic theories such as lexical functional grammar \citep{kaplan1981lexical}, head-driven phrase structure grammar \citep{pollard1994head}, places most of the explanatory weight for morphosyntax on the lexicon -- the properties of the head word (and other words) drive the realization of the rest of the phrase or sentence.
Therefore, we add the lemma for the focus words (e.g.~``dep-lemma-is-cuatro, head-lemma-is-libro'') as  features. 

\paragraph{Semantic Features}
There is  a strong interaction between semantics and sentence structure.
Some well-known examples are of \emph{animacy} or semantic class of a word determining case marking \citep{dahl1996animacy} and word order \cite{thuilier2021word} for some languages.
Continuous vectors \cite{word2vec,bojanowski2017enriching}  have been used to capture semantic (and syntactic) similarity across  words.
However, most vectors are high-dimensional and not easily interpretable, i.e.~what semantic/syntactic property each individual vector value represents is not obvious.
Since our primary goal is to extract comprehensible descriptions of linguistic phenomena, we first generate sparse non-negative vectors using \citet{spine}. 
For each dimension, we extract the top-$k$ words having a high positive value, resulting in features like  dim-1=\{radio,nuclear\}, dim-2=\{hotel,restaurante\}.
This helps us interpret what property each dimension is capturing, for example, dim-1 refers to words about nuclear technology, while dim-2 refers to accommodations.
Now that we can interpret what each feature (dimension) corresponds to, we directly add these vector as features.
In \autoref{fig:systemoverview}, a semantic feature  (e.g.~``dep-word-is-like\!=\!\{ochenta,sesenta\}''\footnote{This translates to \{eight, sixty\}}) extracted for \textbf{cuatro} informs us that the adjective denotes a numeric quantity.

\section{Learning and Extracting Rules}
\label{sec:learning}
\paragraph{Training Data}
To construct the training data $D_{\text{train}}^p$ for each  task $p$,
we start with the raw text $D$ of the language in question and perform syntactic analysis,   producing POS tags, lemmas, morphological analysis and dependency  trees for each sentence. 
Using this analysis, we then identify the focus word(s) and extract $k$ features, forming the input example ($\mathbf{x_i} = \{x_i^0, x_i^1, \cdots, x_i^k\}$).

 \paragraph{Model Training}
Given that the learned model must be interpretable to linguists using the system, we opt to use decision trees \cite{quinlan1986induction}, which split the data into  leaves, where each leaf corresponds to a portion of the input examples following common syntactic/semantic/lexical patterns.

\paragraph{Rule Extraction}
Each leaf in the decision tree is assigned a label based on the distribution of examples within that leaf.
For instance, if a leaf of the adjective-noun word order decision tree has 60\% of examples with adjectives before their nouns, the leaf, by default, is labeled as \emph{before}.
However, a majority-based threshold alone is insufficient as it
does not account for leaves with very few examples, which may be based on spurious correlations or nonsensical feature divisions.
Instead, we use a statistical threshold for leaf labeling, inspired by \citet{chaudhary-etal-2020-automatic}, performing a chi-squared test to first determine which leaves differ significantly from the base distribution. 
For this, we first define the null $H_0$ and test $H_1$ hypotheses. For instance, for word order we define that a leaf:
\begin{align*}
  H_0&: \text{takes either \emph{before/after} label}  \\
  H_1&: \text{takes the label dominant under that leaf}
\end{align*}
We can design such $H_0$ as the words participating in the relation can either be \emph{before} or \emph{after} the other.
To apply the chi-squared test, we compute the expected probability distribution for $H_0$ considering a uniform distribution.
We then compute the p-value and leaves which are not statistically significant are assigned the label of \emph{cannot decide}, which informs a user that the model was uncertain about the label (details in  \autoref{sec:approach}).
Leaves that pass this test are then assigned the majority label and correspond to a rule that will be shown to linguists, where the ``rule''  is described by the syntactic/semantic/lexical features on the branch that lead to that leaf.

\begin{figure*}
\centering
\includegraphics[width=\textwidth]{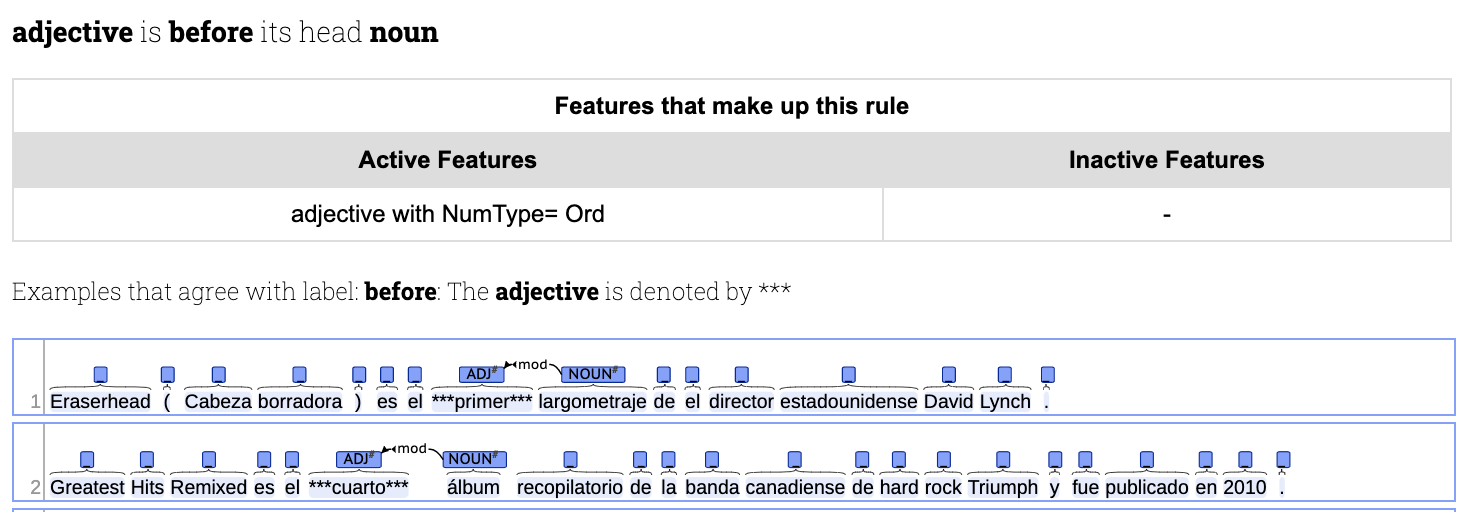}
  \caption{A rule extracted for Spanish adjective-noun word order.  }%
\label{fig:rule}%
\vspace{-1em}
\end{figure*}

\paragraph{Rule Visualization}
For each rule, we extract illustrative examples from the underlying corpus and visualize them in an interface (Figure \ref{fig:rule}).
We select such examples that are both short and consist of diverse word forms to illustrate the rule usage in different contexts.
Along with examples which follow a rule, we also show examples which do not
follow the rule, giving a softer, more nuanced view of the data (details in \autoref{sec:approach}).
Specifically, to not overwhelm the user, we only present 10 examples for each type.

\section{Automated Evaluation Protocol}
In the next two sections, we devise protocols for evaluation of the extracted rules using both automatic metrics (for rapid evaluation that can be applied widely across languages), and evaluation by human language experts (as our gold-standard evaluation).
We first describe below the process of automatic evaluation per linguistic phenomenon.

\paragraph{Case Marking}
As noted earlier, we use the UD scheme for deriving the training data. 
Under this scheme, not every word is labeled with \emph{case}, restricting our training and evaluation to be only on such labeled examples.
For such words, we consider \emph{case} to be a universal property i.e. each word marks a particular \emph{case} value and, we evaluate whether our model can correctly predict that  value.
Thus, we measure the accuracy on a test example $\langle \mathbf{x_i}, y_i \rangle \in  D_{\text{test}}^t$, comparing the models prediction $\hat{y_i}$   with the observed case value $y_i$.
We compare our model against a frequency-based baseline which assigns the most frequent case value in the training data  to all input examples.

\paragraph{Word Order}
Similarly, we can assume that every input example has a word order value, for example subjects will occur either \emph{before} or \emph{after} the verbs.
Therefore, for an input example, we  consider the observed  order to be the ground truth and compute the accuracy by comparing it with the model's prediction.
We compare against a frequency-baseline where the most frequent word order value is assigned to all input examples.


Comparing the model's prediction with the observed order is reasonable for languages which have a dominant word order.
There are a considerable set of languages which have a freer order.
WALS labels such relations as ``no dominant order'' (e.g. subject-verb  order for Modern Greek).
For such cases, considering accuracy alone might be insufficient as there is no  ground truth. 
Therefore, we also report the entropy over the 
predicted  distribution:
\begin{align*}
   H_\text{wo}^r &=  - \sum_{k = \text{before, after}} p_k\log p_k \\
   p_k& =  \frac{\sum_{\langle \mathbf{x_i^r}, y_i \rangle \in  D_{\text{test}}^r}  \mathbbm{1} \left\{
    \begin{array}{ll}
        1 & \hat{y_i} = k \\
        0 & \text{otherwise}
    \end{array}
\right.} {|D_{\text{test}}^r|}
\end{align*}
For languages with no dominant order, the model should be uncertain about the predicted order and we expect the model's entropy to be high. The accuracy computed against the observed  order is still useful, as despite there being ``no dominant order'', speakers tend to prefer one order over the other. A high accuracy would entail that the model was successful in capturing this ``preferred order.''
\paragraph{Agreement}
We use the automated rule metric (ARM) proposed by \citet{chaudhary-etal-2020-automatic} which computes accuracy by comparing the ground truth label to the predicted label.
The ground truth label of an example  is decided using a predefined threshold on the leaf to which the example belongs. 
ARM does not use the observed agreement between the head and its dependent as ground truth because an observed agreement might not necessarily mean \textit{required} agreement.
We compare  with \citet{chaudhary-etal-2020-automatic}, which uses simple syntactic features such as POS of the head, the dependent and, the dependency relation between them.

\begin{table*}[t]
\small
    \centering
    \resizebox{\textwidth}{!}{
    \begin{tabular}{c|c|l|c}
    \toprule
   \textbf{Type} & \textbf{Rule Features} & \textbf{Examples} & \textbf{Label}  \\
    \midrule
    \multirow{2}{*}{Type-1} &  \multirow{4}{*}{Adj is a Ordinal} & \selectlanguage{spanish} También se utilizaba en las \textbf{primeras} \underline{grabaciones} y arreglos jazzísticos. & \multirow{4}{*}{Before} \\
    & &   \textit{It was also used in \textbf{early} jazz \underline{recordings}  and arrangements.}  & \\

     \multirow{2}{*}{(\textcolor{blue}{valid})}& & \selectlanguage{spanish}Las \textbf{primeras} 24 \underline{horas} son cruciales.  \selectlanguage{english} & \\ 
     & &  \textit{The \textbf{first} 24 \underline{hours}  are crucial.}  & \\

   \midrule
   \multirow{2}{*}{Type-2} &  \multirow{2}{*}{Adj belongs to group:} &Matisyahu piensa editar pronto un \textbf{nuevo} \underline{disco} grabado en estudio. \selectlanguage{english} & \multirow{4}{*}{Before}\\
   
   & &  \textit{Matisyahu plans to release a \textbf{new} \underline{studio-recorded} album soon.}  & \\
   
     \multirow{2}{*}{(\textcolor{blue}{valid}, \textcolor{red}{not informative})}& \selectlanguage{spanish} {con,como,no,más,lo} & \selectlanguage{spanish}Es una experiencia \textbf{nueva} \underline{estar} desempleado. \selectlanguage{english} & \\
    & &  \textit{It's a \textbf{new} \underline{experience} being unemployed}  & \\

    \midrule
   \multirow{2}{*}{Type-3} &  \multirow{4}{*}{Adj is NOT Ordinal} &\selectlanguage{spanish}Además de una \textbf{gran} \underline{variedad} de aplicaciones \selectlanguage{english} & \multirow{4}{*}{After}\\
   
   & &  \textit{In addition to a \textbf{great} \underline{variety} of applications.}  & \\
   
     \multirow{2}{*}{(\textcolor{blue}{valid}, \textcolor{orange}{too general})}& & \selectlanguage{spanish}Una \underline{unión} \textbf{solemnizada} en un país extranjero  \selectlanguage{english} & \\
    & &  \textit{An \underline{union} \textbf{solemnized} in a foreign country}  & \\

    \midrule
   \multirow{2}{*}{Type-4} &  \multirow{4}{*}{Adj's lemma is numeroso} &\selectlanguage{spanish}En África hay \textbf{numerosas} \underline{lenguas} tonales  \selectlanguage{english} & \multirow{4}{*}{Before}\\
   
   & &  \textit{In Africa there are \textbf{numerous} tonal \underline{languages}}  & \\
   
     \multirow{2}{*}{(\textcolor{blue}{valid}, \textcolor{orange}{too specific})}& & \selectlanguage{spanish}Ellas poseen   \textbf{varios} \underline{libros}\selectlanguage{english} & \\
    & &  \textit{They own \textbf{several} \underline{books} }  & \\

    \midrule
   \multirow{2}{*}{Type-5} &  \multirow{4}{*}{Adj's head noun is a conjunct} &\selectlanguage{spanish}Las consecuencias  de cualquier (colapso) de divisa e \underline{inflación} \textbf{masiva} . \selectlanguage{english} & \multirow{4}{*}{After}\\
   
   & &  \textit{The consequences expected from any currency collapse and \textbf{massive} \underline{inflation}.}  & \\
   
     \multirow{2}{*}{(\textcolor{red}{invalid})}& & \selectlanguage{spanish}(Realizan) trabajos de alta calidad , muy \textbf{buenos} \underline{profesionales}\selectlanguage{english} & \\
    & &  \textit{They do high quality work, very \textbf{good} \underline{professionals}} & \\

   \bottomrule

    \end{tabular}
    }
    \caption{Types of rules discovered by the model for Spanish adjective-noun word order. \textbf{Adjectives} are highlighted and the \underline{nouns} they modify are underlined. Illustrative examples under each rule are also shown with their English translation in italics.  Label denotes the predicted order. }
    \label{tab:rules}
    \vspace{-1em}
\end{table*}

\section{Human Expert Evaluation Protocol}
\label{sec:experteval}
Since our primary objective is to extract rules which are human-readable and of assistance to the linguists, we  enlist the help of language experts to evaluate the rules on three parameters: \emph{correctness}, \emph{prior knowledge}, \emph{feature correctness}. 
Before starting with the actual evaluation, we first ask the expert to  provide answers regarding the linguistic questions we are evaluating. 
For example, we ask questions such as ``when are subjects after verbs in Greek'', and they are required to provide a brief answer (e.g.~``for questions or when giving emphasis to a subject''). 
We then direct them to our interface where we show the extracted features and a few examples for each rule, then ask questions regarding each of the three parameters (as shown in \autoref{fig:annotation_form} in the Appendix).

Regarding \emph{correctness}, the expert is asked to annotate whether the illustrative examples, shown for that rule, are governed by some underlying grammar rule.
If so, they are then required to judge how precise it is.
Consider some rules extracted for Spanish adjective-noun  order in  \autoref{tab:rules}.
Looking at the examples and features for the Type-1 rule, it is evident that this rule \emph{precisely defines the linguistic distinction}.\footnote{\url{https://www.thoughtco.com/ordinal-numbers-in-spanish-3079591}} 
Some rules, although valid, may be too general (Type-3) or too specific (Type-4). 
Finally, a rule \emph{may not correspond to any underlying grammar rule}, like
the Type-5 where the model simply  discovered a spurious correlation in the data.
For \emph{prior knowledge}, if an extracted rule was indeed a valid grammar rule, then we ask the expert whether they were aware of such a rule. 
This will inform us how useful our framework is in discovering rules which a) align with the expert's prior knowledge and, b) are novel i.e. rules which the expert were not aware of apriori.
Finally, for \emph{feature correctness}, we ask whether the features selected by the model accurately describe said rule. 
For the Type-1 rule, the answer would be \emph{yes}. But for rules like Type-2, the features are not informative even though the corresponding examples do follow a common pattern.



\section{Gold-standard Analysis Experiments}
\label{sec:goldexpt}
In this section, we present results to demonstrate that our framework can discover the conditions  which govern the different linguistic phenomena.
Specifically, we experiment with gold-standard syntactic analysis derived from SUD treebanks, and run experiments to answer questions about word order, agreement, and case marking (\autoref{sec:goldexptauto}).
Furthermore, we manually verify a subset of these extracted rules (\autoref{sec:goldexptmanual}).
Experimenting with languages that have been already studied and have annotated treebanks is crucial for verifying the efficacy of our approach before applying it to other true low- or zero-resource languages. 
Under this setting we not only have clean and expert-annotated data, but we can also quickly compare the effect of data size on the system performance as different languages have treebanks of varying size.

\paragraph{Data and Model}
We use the Syntactic Universal Dependencies v2.5 (SUD) \cite{gerdes-etal-2019-improving} treebanks which are based on the Universal Dependencies (UD) \cite{nivre2016universal,nivre2018universal} project, the difference being that the former allows function words to be syntactic heads (as opposed to UD's preference for content words), which is more conducive to our goal of learning grammar rules.  
We experiment with treebanks for 61 languages, which are publicly available with annotations for POS tags, lemmas, dependency parses, and morphological analysis.
We use the standard SUD train, validation and test splits. 
Syntactic and lexical features are directly extracted from these gold syntactic analyses.
Semantic features are derived from continuous word vectors: we start with 300-dim pre-trained fasttext word vectors~\cite{bojanowski-etal-2017-enriching} which are transformed into sparse vectors using~\citet{spine}\footnote{\url{https://github.com/harsh19/SPINE}}.
Last, we use the \texttt{XGBoost} \cite{chen2016xgboost} library to learn the decision tree. Further details on the model setup are discussed in \autoref{sec:modelsetup}.

\subsection{Automated Evaluation Results}
\label{sec:goldexptauto}
We train models using  syntactic features  for all languages covered by SUD, wherever the linguistic question is applicable.
We find that our models outperform the respective baselines by an (avg.) accuracy of +7.3  for word order, +28.1 for case marking, and +4.0 ARM for agreement.\footnote{We also experimented with Random forests (RF), as suggested by anonymous reviewers, but found the decision trees (DT) to be slightly underperforming ((avg.) -0.12 acc). But given that it is straightforward to extract interpretable rules from DT, which is our primary goal, as compared to RF, we use the former for all experiments, details in Appendix D.}
 We also report the result breakdown under three resource settings, low, mid, and high, where low-resource refers to the treebanks with $<\!500$ sentences, mid-resource has $500\!-\!5000$ sentences and high-resource has $>\!5000$ sentences. 
Across all three linguistic phenomena, the (avg.) model gains over the baseline are +3.19 for the low-resource, +10.7 for the mid-resource and +12.8 for the high-resource. 
The larger the treebank size, the larger the improvement of our model's performance over the baseline.
Even in low-resource settings, a gain over the baseline suggests that our approach is extracting valid rules, which is encouraging for language documentation efforts. 
We present the result breakdown of individual relations in Appendix (\autoref{tab:resultbreakdown}).

\begin{figure}[t]

\begin{tikzpicture}[trim left=-0.8cm,trim right=0cm]
    \begin{axis}[
            ybar,
            every axis plot post/.style={/pgf/number format/fixed},
            bar width=.25cm,
            width=5.0cm,
            height=3cm,
            ymajorgrids=false,
            yminorgrids=false,
            xtick={English, Spanish},
            symbolic x coords={English, Spanish},
            every x tick label/.append style={font=\small},
            every y tick label/.append style={font=\tiny},
            tick pos=left,
            axis x line*=bottom,
            axis y line*=left,
           title={\small \textcolor{blue}{$\blacksquare$} baseline  \textcolor{brown}{$\blacksquare$} syntactic \\ \small \textcolor{teal}{$\blacksquare$} syntactic + lexical  \textcolor{purple}{$\blacksquare$} syntactic + semantic  \\
            \hspace{-4cm} \small Adj-Noun Word Order}, 
            title style={yshift=-0.2cm,xshift=1.8cm,align=center},
            ymin=50,ymax=100,
            ytick={20,40,60,80,100},
            ylabel shift={-.15cm},
            ylabel near ticks,
            ylabel={\small \textsc{accuracy}},
            enlarge x limits=0.6,
        ]
        \addplot [style={blue,fill=blue,bar shift=-.775cm}] coordinates {(English,96.77)};
        \addplot [style={blue,fill=blue,bar shift=-.775cm}] coordinates {(Spanish,68.1)};
       
        \addplot [style={brown,fill=brown,bar shift=-.475cm}] coordinates {(English,	98.25)};
        \addplot [style={brown,fill=brown,bar shift=-.475cm}] coordinates {(Spanish,71.08)};

        \addplot [style={teal,fill=teal,bar shift=-.175cm}] coordinates {(English,98.25)};
        \addplot [style={teal,fill=teal,bar shift=-.175cm}] coordinates {(Spanish,82.65)};
       
        \addplot [style={purple,fill=purple,bar shift=0.175cm}] coordinates {(English,98.25)};
        \addplot [style={purple,fill=purple,bar shift=0.175cm}] coordinates {(Spanish,87.13)};
        
        
    \end{axis}
\end{tikzpicture} \hspace{-.7cm}
\raisebox{0.1cm}{\begin{tikzpicture}[trim left=-4.5cm,trim right=0cm]
    \begin{axis}[
            ybar,
            every axis plot post/.style={/pgf/number format/fixed},
            bar width=.25cm,
            width=5.0cm,
            height=3cm,
            ymajorgrids=false,
            yminorgrids=false,
            xtick={Greek, Turkish},
            symbolic x coords={Greek, Turkish},
            every x tick label/.append style={font=\small},
            every y tick label/.append style={font=\tiny},
            tick pos=left,
            axis x line*=bottom,
            axis y line*=left,
           title={{Noun Case Marking}},
            title style={yshift=-.2cm,font=\small,align=center},
            ymin=40,ymax=100,
            ytick={20,40,60,80,100},
            ylabel shift={-.15cm},
            ylabel near ticks,
            enlarge x limits=0.6,
        ]
        \addplot [style={blue,fill=blue,bar shift=-.775cm}] coordinates {(Greek,49.72)};
        \addplot [style={blue,fill=blue,bar shift=-.775cm}] coordinates {(Turkish,54.65)};
       
        \addplot [style={brown,fill=brown,bar shift=-.475cm}] coordinates {(Greek,	94.87)};
        \addplot [style={brown,fill=brown,bar shift=-.475cm}] coordinates {(Turkish,	64.52)};

        \addplot [style={teal,fill=teal,bar shift=-.175cm}] coordinates {(Greek,94.18)};
        \addplot [style={teal,fill=teal,bar shift=-.175cm}] coordinates {(Turkish,	64.83)};
       
        \addplot [style={purple,fill=purple,bar shift=0.175cm}] coordinates {(Greek,94.96)};
        \addplot [style={purple,fill=purple,bar shift=0.175cm}] coordinates {(Turkish,73.90)};
        
    \end{axis}
\end{tikzpicture}}

\vspace{-1em}
 \caption{Comparing the effect of different features on the word order and  case marking.}
 \label{fig:langfamily}
 \vspace{-1em}
 \end{figure}
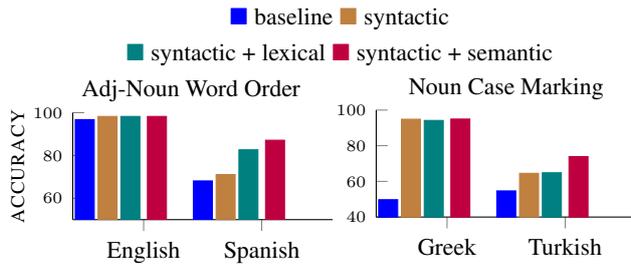

As motivated in \autoref{sec:traindata}, the conditions which govern a linguistic phenomenon vary considerably across languages, which is also reflected in our model's performance.
For example, the model trained on syntactic features alone is sufficient to reach a high accuracy (avg.\!94.2\%) for predicting the adjective-noun  order in Germanic languages. 
 But for Romance languages, using  only syntactic features leads to much lower performance (avg.\!74.6\%).
 We experiment with different features and report results for a subset of languages in \autoref{fig:langfamily}.
Observe that for Spanish adjective-noun order adding lexical features improves the performance significantly (+11.57) over syntactic features, and semantic features provide an additional gain of +4.48.
Studying the languages marked as having ``no dominant order'' in WALS, we find our model does show a higher entropy.
 SUD contains 8 such languages for subject-verb order, and our model produces an (avg.) entropy of 1.09, as opposed to (avg.) 0.75 entropy for all other languages. 
For noun case marking in Greek, syntactic features already bring the model performance to 94\%.
For Turkish, the addition of semantic features raises the model performance by +9.38.
The model now precisely captures that nouns for locations like \selectlanguage{russian}\emph{ev, oda, kapı, dünya}\selectlanguage{english}\footnote{house, room, door, world} typically take the locative case. 
This is in-line with \citet{bamyaci2016animacy} which outlines the importance of animacy in Turkish differential case marking.

 
To confirm that these \emph{discovered conditions generalize to the language as a whole and not the specific dataset on which it was trained}, we train a model on one treebank of a language and apply the trained model directly on the test portions of other treebanks of the same language.
There are 30 languages in the SUD which fit this requirement.
\autoref{fig:treefamily} in the Appendix demonstrates one such setting for understanding the word order 
patterns across different French corpora, where the models have been trained on the largest treebank (\texttt{fr-gsd}).
For subject-verb  order, all treebanks except the \texttt{fr-fqb} show  similar high test performance (~$>$90\% acc.).
Interestingly, the model severely underperforms (28\% acc.) on \texttt{fr-fqb} which is a question-bank corpus comprising of only questions, and questions in French can have varying word order patterns.\footnote{In questions such as \emph{Que \underline{signifie} l' \textbf{acronyme} NASA?} ("What does the acronym NASA mean?"), the \underline{verb} comes before its \textbf{subject}, but for questions such as \emph{\textbf{Qui} \underline{produit} le logiciel ?} ("Who produces the software?") the \textbf{subject} is before the \underline{verb}.}
The model fails to correctly predict the word order because in the training treebank only 1.7\% of examples are questions making it challenging for the model to learn word order rules for different question types.

Through this tool, a linguist can potentially inspect and derive insights on how the patterns discovered for a linguistic question vary across different settings, both within a language and across different languages as well.

\subsection{Human Evaluation Results}
\label{sec:goldexptmanual}
Through the above experiments, we \emph{automatically} evaluated that the extracted rules are predictive (to some extent) and applicable to the language in general. 
Before applying this framework on an endangered language we first perform a manual evaluation ourselves for English and Greek.
We select these languages based on the availability of human annotators, using one expert each for English and Greek. 
First, we note that the total number of rules for English (29) are much less than that for Greek (161), the latter being more morphologically rich. 
We find that 80\% of the rules (across all phenomena) are valid grammar rules for both languages. 
A significant portion  (40\%) of the valid rules are either too specific or too general, which highlights that there is scope of improvement in the feature and/or  model design.
Interestingly, even for English, there were 7 rules which the expert was not aware of. 
For example, the following rule for adjective-noun order -- ``when the nominal is a word like  \emph{something,nothing,anything}, the adjective can come after the noun.``.
For Greek, almost all valid rules were known to the expert, except for one Gender agreement rule\footnote{The rule was, ``proper-nouns modifiers  do not  need to necessarily agree with their head nouns''. }. 
Regarding feature correctness, the Greek expert found 69\% of the valid rules to be readable and informative, while the English expert found 58\% of such rules.
We show the individual results in Appendix (\autoref{fig:q1_results}).

These insights may have utility even for languages that already have automatic NLP tools for POS tagging or dependency parsing, or even a treebank, as existing annotations do not exhaustively describe fine-grained or complex linguistic behaviors on a holistic level (e.g. deviation in word order patterns or explaining the process of agreement). From the user-study above, we do find that the approach discovered fine-grained behaviors for English and Greek, which the language experts were not aware of or could not think of readily. 
In addition, even if language documentation does exist for a language, this does not mean that it is readily available in a standardized machine-readable format, whereas the output of our method is.



\section{Hmong Daw Study}
\label{sec:lowexpt}
Finally, to test the applicability of \textsc{AutoLEX} in a language documentation situation, we experiment with Hmong Daw (mww), a threatened language variety, spoken by roughly 1M people across US, China, Laos, Vietnam and Thailand.
It certainly can be categorized as a low-resourced language with respect to computational resources as well
as accessible and detailed machine-readable grammatical descriptions.
Furthermore, this study presents a realistic setting of language analysis as there is no expert-annotated syntactic analysis available.


We had access to 445k Hmong sentences, which were collected from the \texttt{soc.culture.hmong} Usenet group.
Since the data was scraped from the web, it was noisy and intermixed with English. 
Therefore, first we automatically clean the corpus using a character-level language model trained on English. This automatically filtered 61k sentences.
Next, we automatically obtain syntactic analyses, for which  we train \texttt{Udify} \cite{kondratyuk-straka-2019-75}, a multilingual automatic parser that jointly predicts POS tags, lemmas, morphological analysis and dependency parses, on Vietnamese, Chinese and English treebanks and apply it to the Hmong text.
We randomly split the parsed data into a train and test set (80:20) and apply our general framework to extract rules (details in \autoref{sec:endangeredlang}).


\paragraph{Results}
Hmong has no inflectional morphology so we only train the model to answer word order questions.
We conduct the expert evaluation on four relations where our model outperforms the baseline, albeit slightly (+4.08 for Adj-N, +0.12 for Subj-V, +0.52 for Adp-N, +0.72 for Num-N). For Obj-V relation, our model is on par with the baseline which could indicate that either there were not  many examples whose word order deviated from the dominant order or the  model needs improvement.
First, we ask the expert, a linguist who studies Hmong, to describe the rules (if any) for each relation.
Comparing with the expert's provided rules, we find that the model is successful in discovering the dominant pattern for all relations.
However,  of the 30 rules (across all relations) presented to the expert for annotation, only 5 rules (1 rule for subject-verb, 4 rules for numeral-noun) were found to precisely describe the linguistic distinction. 
For instance, according to the expert, numerals cannot occur immediately before nouns, rather they occur before classifiers which then occur before nouns (``1 clf-1 noun-1'').
Interestingly, one rule captured examples where the numerals were occurring immediately before nouns without the classifiers (e.g. ``1 noun-1, 2 noun-2''), which the expert was not aware of.
On one hand, this is promising as the model, despite being trained on noisy sentences and syntactic analyses, was able to discover instances of interesting linguistic behavior.
However, the expert noted that a large portion of the rules were difficult to evaluate  as these referred to examples which were incorrectly parsed, some of which even described the English portion of code-mixed data.

Despite showing the promise of automatically obtaining
detailed descriptions on languages with good syntactic analyzers, we can see that it is
still challenging to apply methods to such under-resourced languages.
This poses a new challenge for zero-shot parsing, even the relatively strong model of \citet{kondratyuk-straka-2019-75} resulted in a high enough error rate that it impacted the effectiveness of our method, and methods with higher accuracy may further improve the results of end-to-end generation of grammar descriptions.




\section{Next Steps}
While we have demonstrated that our automatic framework can answer linguistic questions across different languages, the rules we discover are limited by the SUD annotation decisions.
For example, several nouns in German are not annotated for the default case, which means these nouns get ignored by our model in the current setting.
Possibly, using language-specific annotations or heuristics could help alleviate this problem.
As noted in the Hmong study, the quality of rules depends on the quality of the underlying parses. 
We plan to devise an iterative process where a linguist, assisted by an automatic parser, can improve syntactic parsing. The model extracts rules using improved analyses, which the linguist can  inspect and provide more inputs to further improve.
We note that currently we use English as the meta-language to describe the rules, which assumes that an \textsc{AutoLEX} user is well-versed with English and the corresponding grammar terms.
In the future, we plan to provide these rules in the user's choice of language.

\section*{Statement of Ethics}
We acknowledge that there are several ethical concerns while working with endangered or threatened languages, in particular, that we include and take guidance from community members when designing any technology using their data. Secondly, that any data we collect is appropriately used, without causing any detrimental effect or bias on the community. 
In adherence to that, this work is done in collaboration with a Hmong linguist who is in close collaboration and consultation with the community. 
We will release any tools that we build for Hmong in consultation with them and the community. 

\section*{Acknowledgements}
The authors are grateful to the anonymous reviewers who took the time to provide many interesting
comments that made the paper significantly better.
This work is sponsored by the
Waibel Presidential Fellowship and by the National
Science Foundation under grant 1761548.

\bibliographystyle{acl_natbib}
\bibliography{acl}

\clearpage
\newpage

\appendix
\newpage
\appendix

\section{Related Work}
\label{sec:relatedwork}
Prior work  \cite{lewis-xia-2008-automatically, hellan-2010-descriptive,bender-etal-2013-towards,howell-etal-2017-inferring} have proposed methods to map descriptive grammars, present in the form of inter-linear glossed text (IGT), to existing head-phrase structure grammar (HPSG) based grammar system which is machine-readable.
\citet{lewis-xia-2008-automatically} enrich IGT data with syntactic structures to determine canonical word order and case marking observed in the language.
They do note that, while a linguist carefully chooses the examples to create the IGT corpus such that they are representative of the linguistic phenomena of interest, insights derived from IGT may suffer from this bias as the data doesn't encompass many of the naturally-occurring examples.
\citet{hellan-2010-descriptive} present a sentence-level annotation code which maps the properties of the sentence to discrete labels. 
These discrete labels form a template which are then mapped to in a mixed to HPSG or LFG format \cite{pollard1994head, kaplan1981lexical}.
\citet{bender-etal-2013-towards} extract major-constituent word order and case marking properties from the IGT for a diverse set of languages.
Potentially, grammar rules can also be derived from existing projects such as the LinGO Grammar Matrix \cite{bender-etal-2002-grammar}, ParGram \cite{butt-etal-2002-parallel,king2005feature}. 
These are grammar development tools designed to write and create grammar specifications that support a wide range of languages, in a unified format.
They focus on mapping simple description of languages, obtained from existing IGT-annotated data or input from a linguist, to precision grammar fragments, grounded in a grammar formalism such as HPSG, LFG.
Our work differs in that, 1) we attempt to discover and explain the local linguistic behaviors for the language in general, 2) we do not extract rules for an  individual sentence in isolation, as some of the HPSG/LFG-based approaches do, 3) we discover these behaviors from naturally occurring sentences.
We do note that the rules we present in this work are based on the SUD annotation scheme, but the current framework can be easily extended to any other such scheme.
In \autoref{tab:relatedwork}, we outline the different linguistic questions answered by our work and the related work.
\begin{table*}[t]
\small
    \centering
    \resizebox{\textwidth}{!}{
    \begin{tabular}{c|l|c|c}
    \toprule
  Linguistic Phenomena &  Work & Rule-Type & Corpus Type   \\
    \midrule
  WordOrder & Ours & C+FG & Raw text\\
            & Grammar Matrix \cite{bender-etal-2002-grammar} & C+FG & IGT text* \\
            & \citet{lewis-xia-2008-automatically} & C &  IGT text \\
            & \citet{bender-etal-2013-towards} & C & IGT text \\
            & \citet{ostling-2015-word} & C & Raw text   \\
            & \citet{wang-eisner-2017-fine} & C & Raw text  \\
            & WALS \citet{wals} & C & Reference grammar* \\
  \midrule
  Case Marking & Ours & C+FG & Raw text \\
               & Grammar Matrix \cite{bender-etal-2002-grammar} & C+FG & IGT text* \\
               & WALS \citet{wals} & C & Reference grammar* \\
               & \citet{howell-etal-2017-inferring} & C & IGT text \\
    \midrule
    Agreement &  Ours & C+FG & Raw text \\
              & Grammar Matrix \cite{bender-etal-2002-grammar} & C+FG & IGT text* \\
              & \citet{chaudhary-etal-2020-automatic} & C+FG & Raw text \\
              
    \midrule 
    Sentence construction & \citet{hellan-2010-descriptive} & FG & IGT text* \\
   
    \end{tabular}
    }
    \caption{An overview of linguistic questions \emph{automatically} answered by our current work and existing related work. Some of them combine semi-automatic approaches with manually annotated resources, there are marked with *. Rule-Type  denotes the type of rule extracted for a language, C refers to coarse-grained  such as rules for canonical word order, FG refers to fine-grained i.e. rules extracted at a local level. 
    }
    \label{tab:relatedwork}
    \vspace{-1em}
\end{table*}

There has also been work on developing toolkits to visualize some aspects of language structure -- \citet{lepp-etal-2019-visualizing} present a web-based system to explore different morphological analyses. They also allow a user to improve the analyses thereby also improving the grammar specification which relies on those analyses.

\section{Learning and Extracting Rules}
\label{sec:approach}
\paragraph{Statistical Threshold for Rule Extraction}
Similar to \citet{chaudhary-etal-2020-automatic}, we apply statistical testing to label leaves. For morphological agreement, we use the same hypothesis definition where the null hypothesis $H_0$ states that \emph{each leaf denotes chance-agreement}.
This means that there is no required agreement between a head and its dependent on the morphological attribute $m$.
The hypothesis to be tested for is $H_1$ which states that \emph{the leaf denotes required-agreement}.
For case marking, we follow a similar approach as explained for word order. We can design $H_0$ as word order,  because under the abstract case viewpoint (\autoref{sec:traindata}), case is a universal property for each word.
We use a $\text{p}-\text{value}= 0.01$ based on the recommendation of \citet{chaudhary-etal-2020-automatic}.

\paragraph{Rule Visualization}
Under each rule, we present a subset of examples from the training portion of the treebank to illustrate the rule.
Positive examples refer to the examples which have  features (from that rule)
and follow the label as predicted by that leaf.
However, there could be examples in the training data which have the same features as defined under that rule, but these example do not follow the predicted label. We refer to these examples as  negative examples.

Since we only show a small set of examples, we select these examples to be concise and representative.
We first group the examples under the rule with the lemmatized forms of the focus words.
For example, under the Type-1 rule (\autoref{tab:rules}) extracted for Spanish adjective-noun word order, the focus words are the \textbf{adjective} ($w_a$) and the \underline{noun} ($w_b$). We group these examples by the lemmatized forms of the adjective and noun $\langle l_a, l_b \rangle$. 
The examples grouped under a lemmatized pair $\langle l_a, l_b \rangle$ are then sorted by their  lengths.
For each lemmatized pair $\langle l_a, l_b \rangle$, we select the top shortest examples.
Finally, all selected examples are shuffled and we randomly select 10 examples.

\section{Experimental Setup}
\label{sec:modelsetup}
\paragraph{Data}
Below we describe the license details of the datasets we used:
\begin{itemize}
    \item \texttt{SUD treebanks}: No specific license is specified, but the data is released as part of research work \cite{gerdes-etal-2019-improving}. We have used this data as intended which is for academic research purposes.
    
    \item \texttt{Fasttext embeddings}: Released\footnote{\url{https://fasttext.cc/docs/en/pretrained-vectors.html}} under the  Creative Commons Attribution-Share-Alike License 3.0.
    We have used this data as intended, which is for academic research purposes.
    
    \item \texttt{Hmong Daw}: This dataset was collected by one of the co-authors from the Usenet group \texttt{soc.culture.hmong} and is currently in submission to LREC. 
    The data used will be released as part of the Creative Commons Zero v1.0 Universal license.
    Accordingly, we will also release the train/test split for better reproducibility. 
    
    The obvious identifying information has been removed from the data, although it would be possible to recover that information by going back to the original Usenet posts.
\end{itemize}
\paragraph{Model}
As described in the main text, we use the \texttt{XGBOOST} to learn a decision tree.
For each language, the running time of the model is  approximately 2-5 mins.
We perform a grid search over a set of hyperparameters and select the best performing  model based on the validation set performance.
Here the hyperparameters we use:
\begin{itemize}
    \item \texttt{criterion}: \{gini, entropy\}
    \item \texttt{max-depth}: \{3, 4, 5, 6, 7, 8, 9, 10, 15, 20\}
    \item \texttt{n-estimators}: 1
    \item \texttt{learning-rate}: 0.1
    \item \texttt{objective}: multi:softprob
\end{itemize}

\section{Gold-standard Experiments}
\subsection{Automated Evaluation Results}
In the main text, we reported the average improvement for the word order, agreement and case marking models. In  \autoref{tab:resultbreakdown} we present the breakdown per each  question.
The word order results are reported over 56 languages, agreement over 38 and  case marking over 35 languages.\footnote{Some languages have very little training data on which we couldn't fit a model while for some languages the linguistic questions was not applicable.}
\begin{table}[t]
\small
    \centering
    \resizebox{\columnwidth}{!}{
    \begin{tabular}{c|l|c}
   Linguistic Phenomena & Model & Gain  \\
    \midrule
    Word Order & adjective-noun &  2.61\\
               & subject-verb & 6.95\\
               & object-verb & 10.78\\
               & numeral-noun &9.88 \\
               & noun-adposition & 2.31\\
    \midrule
    Agreement & Gender &  4.02 \\
              & Person & 1.08\\
              & Number & 4.95 \\
    \midrule
    Case Marking & NOUN & 30.03\\
                 & PRON & 32.66 \\
                & DET & 47.33 \\
                & PROPN & 29.77 \\    
                & ADJ & 35.59 \\
                & VERB & 18.76 \\
                & ADP  & 15.4 \\
                & NUM  & 25.81 \\
    \end{tabular}
    }
    \caption{Breakdown of the performance gain (over the baseline) for each linguistic question. The performance of the agreement models is compared with the models trained over simple syntactic features in \citet{chaudhary-etal-2020-automatic}. }
    \label{tab:resultbreakdown}
    \vspace{-1em}
\end{table}

\begin{table}[t]
\small
    \centering
    \resizebox{\columnwidth}{!}{
    \begin{tabular}{c|l|c|c|c}
   Model  & Language & Random forest (acc.) & Decision tree (acc.) & Baseline  \\
   \midrule
adjective-noun & el & 99.29 & 99.29 & 99.29 \\
 & es & \textbf{73.32} & 71.46 & 68.1 \\
 & ur & 99.04 & 99.04 & 99.04 \\
 & fi & 98.37 & \textbf{99.09} & 98.37 \\
 & lv & 98.84 & 98.84 & 98.84 \\
 & it & \textbf{70.4} & 69.26 & 66.02 \\
 & no & 97.92 & 97.92 & 97.76 \\
 & fr & \textbf{74.01} & 73.6 & 73.6 \\
 & ro & \textbf{95.83} & 95.19 & 92.95 \\
 & bg & 97.98 & \textbf{98.49} & 97.23 \\
 & gl & 79.2 & 79.2 & 79.2 \\
 \midrule
subject-verb & en & 98.81 & 98.81 & 94.15 \\
 & el & \textbf{85.52} & 83.45 & 73.56 \\
 & es & \textbf{83.5} & 82.52 & 71.52 \\
 & tr & 92.96 & 92.96 & 92.96 \\
 & hi & 99.56 & 99.56 & 99.56 \\
 & fi & 87.14 & \textbf{90.36} & 79.16 \\
 & lv & \textbf{79.79} & 77.73 & 73.99 \\
 & it & \textbf{82.37} & 81.44 & 71.76 \\
 & no & \textbf{86.28} & 85.33 & 70.34 \\
 & fr & 94.21 & 94.21 & 94.21 \\
 & ug & 95.13 & 95.13 & 95.13 \\
 & ro & \textbf{75.62} & 73.49 & 54.36 \\
 & bg & \textbf{81.67} & 79.22 & 72.73 \\
 & gl & \textbf{86.26} & 85.5 & 82.14 \\
 \midrule
object-verb & en & \textbf{98.8} & 98.66 & 97.26 \\
 & el & 96.2 & 96.2 & 86.0 \\
 & es & 95.99 & 95.99 & 90.4 \\
 & tr & 96.64 & 96.64 & 96.64 \\
 & hi & \textbf{99.78} & 99.61 & 74.71 \\
 & ur & 99.45 & \textbf{99.59} & 79.5 \\
 & fi & 85.21 & \textbf{86.36} & 74.83 \\
 & lv & \textbf{83.31} & 82.95 & 75.24 \\
 & it & \textbf{94.97} & 94.79 & 84.92 \\
 & no & 98.68 & 98.68 & 95.86 \\
 & fr & \textbf{96.96} & 96.53 & 86.33 \\
 & ro & 86.99 & \textbf{87.79} & 65.06 \\
 & bg & \textbf{92.53} & 92.22 & 80.66 \\
 & gl & \textbf{94.48} & 94.17 & 82.2 \\
 \midrule
noun-adposition & en & 99.42 & 99.42 & 99.42 \\
 & es & 100.0 & 100.0 & 98.83 \\
 & ur & 98.91 & 98.91 & 98.91 \\
 & fi & 89.35 & \textbf{98.12} & 89.35 \\
 & lv & 97.78 & 97.78 & 97.78 \\
 & no & \textbf{99.3} & 99.26 & 99.14 \\
 & gl & 99.32 & 99.32 & 99.18 \\
 \midrule
numeral-noun & en & 88.06 & 88.06 & 82.09 \\
 & el & 80.6 & 80.6 & 80.6 \\
 & es & 88.62 & 88.62 & 75.61 \\
 & ur & 95.63 & 95.63 & 95.63 \\
 & fi & \textbf{92.14} & 87.25 & 90.71 \\
 & it & \textbf{82.33} & 79.32 & 79.32 \\
 & no & 85.78 & \textbf{88.44} & 88.44 \\
 & fr & 81.16 & \textbf{81.88} & 60.87 \\
 & ro & 84.14 & \textbf{84.83} & 62.07 \\
 & bg & 88.24 & 88.24 & 88.24 \\
    \end{tabular}
    }
    \caption{Comparing the accuracy of Random forest classifier with the Decision Tree for different word order relations. }
    \label{tab:woresult}
    \vspace{-1em}
\end{table}

We also report results under three resource settings as shown in \autoref{tab:resultbreakdownresource}.
 \begin{table}[t]
\small
    \centering
    \resizebox{\columnwidth}{!}{
    \begin{tabular}{c|l|c}
   Linguistic Phenomena & Resource-Setting  & Gain (number of models)  \\
    \midrule
    Agreement & low &  -3.39 (10) \\
              & mid & 1.07 (25)\\
              & high & 5.89 (55) \\
    \midrule
    Case Marking & low & 12.14 (11)\\
                 & mid & 28.17 (56)\\
                & high & 37.17 (56)\\
    \end{tabular}
    }
    \caption{Breakdown of the performance gain (over the baseline) for each linguistic question by resource setting.  }
    \label{tab:resultbreakdownresource}
    \vspace{-1em}
\end{table}

We also show individual results per each language for word order (\autoref{tab:wordorderall-1}, \autoref{tab:wordorderall-2}), agreement (\autoref{tab:agreementall-1}), case marking (\autoref{tab:assignmentall-1}, \autoref{tab:assignmentall-2}).
We note that we report these results on a single run of the experiment.
\begin{figure}[t]

\pgfplotstableread[row sep=\\,col sep=&]{
Type & fr-gsd & fr-partut & fr-pud & fr-fqb & fr-ftb & fr-spoken & base-fr-gsd & base-fr-partut & base-fr-pud & base-fr-fqb & base-fr-ftb & base-fr-spoken \\
Adj-N & 89.81 & 75.56 &	82.22 & 88.23 & 30.33 & 82.21 & 73.6 & 60  &67.78 & 70.73 & 69.28 & 	54.91\\
Obj-V & 97.61 &	97.7 &	99.27 &	79.93 &	96.86 &	95.18 & 	86.33 &	91.95 &	82.79 &	70.45 &	88.29 &	73.67 \\
Subj-V & 96.04 &	92.31 &	95.78 &	28.18 &	92.86 &	97.43 & 94.21 &	95.38 &	95.78 &	28.18 &	92.86 &	97.43	\\
}\zeroshotdata

\begin{tikzpicture}[trim left=-0.6cm,trim right=0cm]
    \begin{axis}[
            ybar,
            every axis plot post/.style={/pgf/number format/fixed},
           bar width=.05cm,
            width=8.5cm,
            height=4cm,
            ymajorgrids=false,
            yminorgrids=false,
            xtick={Subj-V,Adj-N, Obj-V},
            symbolic x coords={Subj-V,Adj-N, Obj-V},
            every x tick label/.append style={font=\small},
            every y tick label/.append style={font=\tiny},
            tick pos=left,
            axis x line*=bottom,
            axis y line*=left,
            title={\small \textcolor{blue}{$\blacksquare$} fr-gsd  \textcolor{orange}{$\blacksquare$} fr-partut
            \textcolor{teal}{$\blacksquare$} fr-pud
            \textcolor{red}{$\blacksquare$} fr-ftb  
             \textcolor{brown}{$\blacksquare$} fr-fqb
            \textcolor{purple}{$\blacksquare$} fr-spoken},  
            title style={yshift=0cm,align=left},
            ymin=20,ymax=100,
            ytick={20,40,60,80,100},
            ylabel shift={-.15cm},
            ylabel near ticks,
            ylabel={\small \textsc{accuracy}},
            enlarge x limits=0.2,
        ]
        \addplot [style={blue,fill=blue,opacity=0.4,bar shift=-0.3cm, bar width=.15cm}] table[x=Type,y=fr-gsd]{\zeroshotdata};
        \addplot [style={blue,fill=blue,bar shift=-0.3cm}] table[x=Type,y=base-fr-gsd]{\zeroshotdata};
        \addplot [style={orange,fill=orange,opacity=0.4,bar shift=-0.15cm, bar width=.15cm}] table[x=Type,y=fr-partut]{\zeroshotdata};
        \addplot [style={orange,fill=orange,bar shift=-0.15cm}] table[x=Type,y=base-fr-partut]{\zeroshotdata};
         \addplot [style={teal,fill=teal,opacity=0.4,bar shift=0cm, bar width=.15cm}] table[x=Type,y=fr-pud]{\zeroshotdata};
        \addplot [style={teal,fill=teal,bar shift=0cm}] table[x=Type,y=base-fr-pud]{\zeroshotdata};

        \addplot [style={red,fill=red,opacity=0.4,bar shift=0.15cm, bar width=.15cm}] table[x=Type,y=fr-ftb]{\zeroshotdata};
        \addplot [style={red,fill=red,bar shift=0.15cm}] table[x=Type,y=base-fr-ftb]{\zeroshotdata};
        
         \addplot [style={brown,fill=brown,opacity=0.4,bar shift=0.3cm, bar width=.15cm}] table[x=Type,y=fr-fqb]{\zeroshotdata};
        \addplot [style={brown,fill=brown,bar shift=0.3cm}] table[x=Type,y=base-fr-fqb]{\zeroshotdata};

         \addplot [style={purple,fill=purple,opacity=0.4,bar shift=0.45cm, bar width=.15cm}] table[x=Type,y=fr-spoken]{\zeroshotdata};
        \addplot [style={purple,fill=purple,bar shift=0.45cm}] table[x=Type,y=base-fr-spoken]{\zeroshotdata};
    
    \end{axis}
\end{tikzpicture}
 \caption{Comparing the accuracy of the model across different treebanks. Each model is trained on the fr-gsd treebank and directly applied on the other treebanks. Shaded bars denote the best model performance trained using all features while solid bars denote the most-frequent baseline for that treebank. }
 \label{fig:treefamily}
 \end{figure}
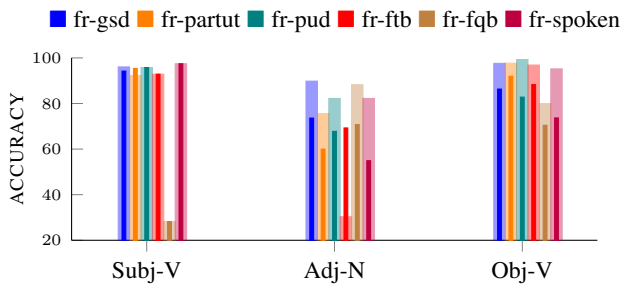

 We experiment with Random Forest, which is a better classifier, in comparison to decision trees, but it is not as interpretable as the latter. 
 Nevertheless, as requested by the anonymous reviewers, we compare how decision trees fare against Random forest in \autoref{tab:woresult}. 
 We train models for answering word order questions, across 15 languages from the SUD treebanks.
 Overall, we observe that decision trees slightly underperform the Random forest, but by only (avg.) -0.12 acc. points, where the range of accuracy is 0-100.
 Given our primary goal is to extract comprehensible descriptions, we opt to use decision trees. 
 
\subsection{Human Evaluation Results}
\label{sec:engresults}
\input{images/human-results}
We conduct expert evaluation for English and Greek. 
Both the English and Greek language expert are co-authors of the paper. 
For English, a total of 15 rules were evaluated for agreement, 11 for word order and 3 for case marking.
For Greek, a total of  35 rules were evaluated for agreement, 11 for word order and 115  for case marking.
We discussed the results in the main text, here we present the figures for English and Greek (\autoref{fig:q1_results}).
For English, there were some rules which the expert was not aware of. We discussed one example for word order in the main text, 
we show an example for agreement and case marking in \autoref{tab:english-rules}.

\begin{table*}[t]
\small
    \centering
    \resizebox{\textwidth}{!}{
    \begin{tabular}{c|c|l|c}
    \toprule
   Linguistic Phenomena & Rule & Examples & Label  \\
    \midrule
    Number & dependent's head is a NOUN &  
        \textbf{Kids} fun \underline{games} are added to the building. & Not-required-agreement \\

        Agreement & &
        \textbf{Nationalist} \underline{groups} are coming to the conference. &  \\
        
        \midrule
    Object & Pronoun is a oblique &  
        Because Large Fries give \textbf{you} FOUR PIECES ! & Accusative \\

        Case Marking & &
        Give \textbf{him} a call tommorow &  \\

    \end{tabular}
    }
    \caption{Some example of rules for agreement and case marking, which the expert annotator was not aware of.
    The 
    \textbf{focus word} is highlighted, for agreement we also underline the \underline{head} with which the dependent's agreement is checked.
    The examples under number agreement demonstrate that when dependent's head is a noun the \textbf{dependent} need not agree with its \underline{head}. We show one example where the first example shows the dependent matches the number  of the head, and the second example shows that it didn't not match.   }
    \label{tab:english-rules}
    \vspace{-1em}
\end{table*}

\begin{figure*}
    \centering
    \includegraphics[width=\textwidth]{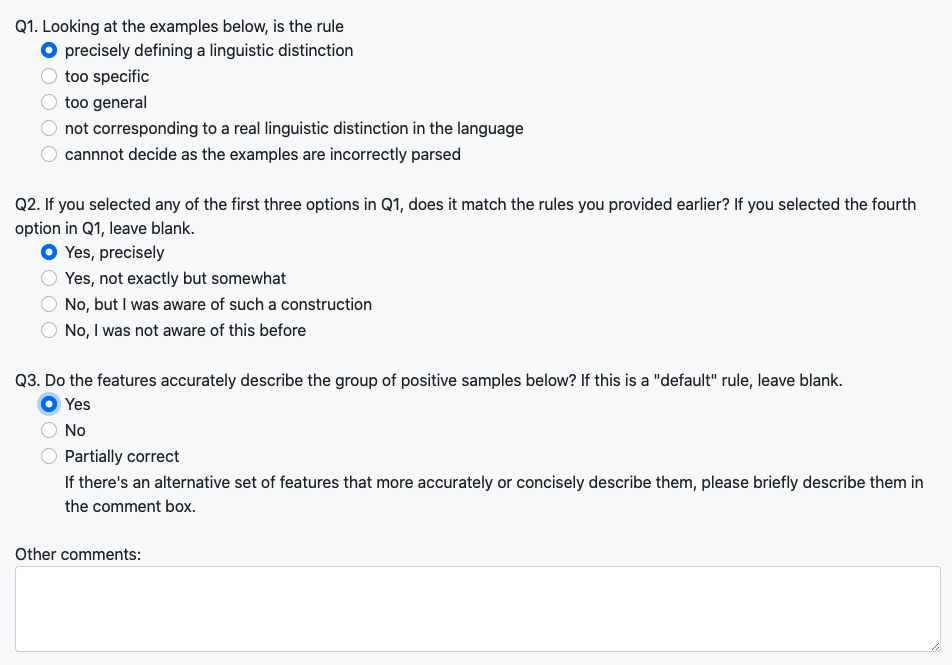}
    \caption{Rule evaluation form presented to the language expert. }
    \label{fig:annotation_form}
\end{figure*}

\section{Hmong Daw Study}
\label{sec:endangeredlang}
\paragraph{Data}
We experimented with the Hmong Daw variety in this setting.
One of co-authors of the paper is a Hmong linguist who is in close
collaboration and consultation with the community, and is the expert who provided us with the Hmong data and also helped evaluate the extracted grammar rules.
We chose Vietnamese, Chinese and English to train \texttt{udify} model as they share syntactic and lexical similarity with Hmong.
We use the same hyperparameter setting as specified in the code\footnote{\url{https://github.com/Hyperparticle/udify}}.


\begin{table*}[t]
\small
    \centering
    \resizebox{\textwidth}{!}{
    \begin{tabular}{c|c|c||c|c|c}
    \toprule
    Type & Lang & Train - Test - Baseline & Type & Lang & Train - Test - Baseline  \\
    \midrule

 adjective-noun & it-vit & 70.71 - \textbf{69.51} - 66.02  & object-verb & cu-proiel & 80.37 - \textbf{82.72} - 76.03  \\
adjective-noun & no-nynorsk & 97.68 - \textbf{97.92} - 97.76  & object-verb & be-hse & 87.79 - 95.38 - \textbf{95.38}  \\
adjective-noun & ro-nonstandard & 87.46 - \textbf{95.19} - 92.95  & object-verb & sv-lines & 96.75 - \textbf{96.79} - 95.31  \\
adjective-noun & bg-btb & 97.27 - \textbf{98.49} - 97.23  & object-verb & uk-iu & 82.77 - \textbf{87.16} - 83.16  \\
adjective-noun & gl-ctg & 79.02 - 79.2 - \textbf{79.2}  & object-verb & ga-idt & 94.89 - \textbf{91.55} - 82.8  \\
adjective-noun & cs-pdt & 94.69 - \textbf{94.36} - 93.69  & object-verb & sk-snk & 81.84 - \textbf{86.17} - 80.91  \\
adjective-noun & fi-tdt & 98.56 - 99.09 - \textbf{99.09}  & object-verb & hu-szeged & 73.23 - \textbf{68.26} - 53.73  \\
adjective-noun & pl-pdb & 65.61 - \textbf{68.0} - 61.84  & object-verb & got-proiel & 74.58 - \textbf{80.15} - 72.44  \\
adjective-noun & la-ittb & 63.64 - \textbf{59.65} - 40.2  & object-verb & hr-set & 89.27 - \textbf{92.2} - 83.32  \\
adjective-noun & nl-alpino & 98.38 - 98.65 - \textbf{98.65}  & object-verb & lzh-kyoto & 97.86 - \textbf{98.01} - 95.7  \\
adjective-noun & mt-mudt & 78.91 - 82.84 - \textbf{82.84}  & object-verb & lv-lvtb & 85.03 - \textbf{82.95} - 75.24  \\
adjective-noun & ja-bccwj & 99.4 - 98.69 - \textbf{98.69}  & object-verb & et-edt & 76.03 - \textbf{79.51} - 69.67  \\
adjective-noun & orv-torot & 71.39 - \textbf{65.76} - 53.48  & object-verb & fro-srcmf & 79.62 - \textbf{81.82} - 48.25  \\
adjective-noun & pt-gsd & 70.31 - \textbf{74.54} - 71.63  & object-verb & af-afribooms & 82.72 - \textbf{96.19} - 86.03  \\
adjective-noun & cu-proiel & 84.96 - 84.98 - \textbf{84.98}  & object-verb & hy-armtdp & 71.47 - \textbf{74.58} - 44.92  \\
adjective-noun & sv-lines & 98.3 - \textbf{98.29} - 95.67  & object-verb & en-ewt & 98.33 - \textbf{98.94} - 97.26  \\
adjective-noun & uk-iu & 94.68 - 95.19 - \textbf{95.19}  & object-verb & fr-gsd & 98.89 - \textbf{97.18} - 86.33  \\
adjective-noun & sk-snk & 96.11 - 95.17 - \textbf{95.17}  & object-verb & el-gdt & 97.18 - \textbf{96.2} - 86.0  \\
adjective-noun & got-proiel & 79.51 - \textbf{79.51} - 72.48  & object-verb & es-gsd & 97.47 - \textbf{95.99} - 90.4  \\
adjective-noun & hr-set & 96.24 - \textbf{96.78} - 96.36  & object-verb & tr-imst & 95.38 - 96.64 - \textbf{96.64}  \\
adjective-noun & lv-lvtb & 98.93 - 98.84 - \textbf{98.84}  & object-verb & ru-syntagrus & 87.47 - \textbf{88.33} - 85.63  \\
adjective-noun & et-edt & 99.57 - \textbf{99.36} - 99.01  & object-verb & sl-ssj & 84.16 - \textbf{88.24} - 72.92  \\
adjective-noun & fro-srcmf & 73.84 - \textbf{74.42} - 73.26  & object-verb & id-gsd & 99.33 - \textbf{98.99} - 95.97  \\
adjective-noun & en-ewt & 97.84 - \textbf{98.25} - 96.77  & object-verb & lt-alksnis & 80.76 - \textbf{79.02} - 69.73  \\
adjective-noun & fr-gsd & 71.04 - \textbf{73.8} - 73.6  & object-verb & ar-nyuad & 96.27 - \textbf{95.91} - 95.63  \\
adjective-noun & el-gdt & 97.34 - 99.29 - \textbf{99.29}  & object-verb & grc-proiel & 72.98 - \textbf{75.87} - 67.05  \\
adjective-noun & es-gsd & 76.27 - \textbf{71.46} - 68.1  & subject-verb & it-vit & 82.95 - \textbf{82.53} - 71.76  \\
adjective-noun & ru-syntagrus & 97.84 - \textbf{98.0} - 96.54  & subject-verb & no-nynorsk & 83.42 - \textbf{85.33} - 70.34  \\
adjective-noun & sl-ssj & 98.22 - \textbf{98.27} - 97.78  & subject-verb & ug-udt & 95.32 - 95.13 - \textbf{95.13}  \\
adjective-noun & id-gsd & 93.41 - 92.79 - \textbf{92.79}  & subject-verb & ro-nonstandard & 69.06 - \textbf{74.27} - 54.36  \\
adjective-noun & lt-alksnis & 98.61 - 98.3 - \textbf{98.3}  & subject-verb & bg-btb & 78.86 - \textbf{79.65} - 72.73  \\
adjective-noun & ar-nyuad & 99.65 - 99.64 - \textbf{99.64}  & subject-verb & gl-ctg & 84.54 - \textbf{85.5} - 82.14  \\
adjective-noun & grc-proiel & 65.23 - \textbf{72.33} - 64.82  & subject-verb & cs-pdt & 67.13 - \textbf{73.18} - 63.33  \\
adjective-noun & de-hdt & 99.47 - \textbf{99.66} - 99.26  & subject-verb & fi-tdt & 88.11 - \textbf{90.57} - 88.19  \\
object-verb & it-vit & 96.28 - \textbf{94.88} - 84.92  & subject-verb & pl-pdb & 78.19 - \textbf{80.6} - 72.1  \\
object-verb & no-nynorsk & 97.73 - \textbf{98.68} - 95.86  & subject-verb & la-ittb & 80.29 - \textbf{82.69} - 72.54  \\
object-verb & ro-nonstandard & 86.05 - \textbf{87.79} - 65.06  & subject-verb & zh-gsd & 99.78 - \textbf{99.44} - 97.39  \\
object-verb & bg-btb & 92.18 - \textbf{92.43} - 80.66  & subject-verb & nl-alpino & 70.62 - \textbf{72.11} - 67.12  \\
object-verb & gl-ctg & 92.71 - \textbf{94.17} - 82.2  & subject-verb & mt-mudt & 83.91 - \textbf{84.96} - 72.03  \\
object-verb & cs-pdt & 82.35 - \textbf{83.91} - 73.97  & subject-verb & orv-torot & 72.38 - \textbf{66.07} - 60.46  \\
object-verb & fi-tdt & 84.21 - \textbf{86.62} - 77.98  & subject-verb & he-htb & 73.43 - \textbf{70.7} - 63.44  \\
object-verb & pl-pdb & 88.89 - \textbf{90.28} - 81.07  & subject-verb & pt-gsd & 89.4 - \textbf{93.15} - 87.47  \\
object-verb & la-ittb & 65.96 - \textbf{65.36} - 52.63  & subject-verb & cu-proiel & 73.88 - \textbf{76.31} - 62.48  \\
object-verb & zh-gsd & 93.4 - \textbf{94.12} - 87.75  & subject-verb & be-hse & 82.86 - \textbf{83.33} - 81.11  \\
object-verb & nl-alpino & 90.32 - \textbf{94.69} - 47.48  & subject-verb & sv-lines & 80.17 - \textbf{80.72} - 73.06  \\
object-verb & mt-mudt & 95.66 - 94.96 - \textbf{94.96}  & subject-verb & uk-iu & 76.89 - \textbf{77.14} - 74.56  \\
object-verb & wo-wtb & 91.6 - \textbf{91.81} - 75.11  & subject-verb & ga-idt & 99.33 - \textbf{99.28} - 85.25  \\
object-verb & orv-torot & 76.71 - \textbf{72.56} - 65.51  & subject-verb & sk-snk & 63.43 - 73.69 - \textbf{73.69}  \\
object-verb & he-htb & 97.87 - 98.03 - \textbf{98.03}  & subject-verb & hu-szeged & 75.91 - \textbf{74.59} - 72.43  \\
object-verb & pt-gsd & 95.17 - \textbf{95.02} - 88.45  & subject-verb & got-proiel & 67.56 - \textbf{73.2} - 66.17  \\
  \end{tabular}
    }
    \caption{Accuracy results for all relations across different languages. Baseline is the most frequent order in the training data.}
    \label{tab:wordorderall-1}
    \vspace{-1em}
\end{table*}

\begin{table*}[t]
\small
    \centering
    \resizebox{\textwidth}{!}{
    \begin{tabular}{c|c|c||c|c|c}
    \toprule
    Type & Lang & Train - Test - Baseline & Type & Lang & Train - Test - Baseline  \\
    \midrule 

subject-verb & hr-set & 81.87 - \textbf{86.62} - 77.44  & noun-adposition & fi-tdt & 97.88 - \textbf{98.12} - 89.47  \\
subject-verb & cop-scriptorium & 85.92 - \textbf{83.84} - 76.71  & noun-adposition & pl-pdb & 99.97 - \textbf{99.97} - 99.83  \\
subject-verb & lv-lvtb & 76.96 - \textbf{77.98} - 73.99  & noun-adposition & nl-alpino & 99.28 - \textbf{99.57} - 99.23  \\
subject-verb & et-edt & 68.13 - \textbf{71.93} - 61.02  & noun-adposition & orv-torot & 97.92 - \textbf{97.54} - 96.83  \\
subject-verb & fro-srcmf & 79.21 - \textbf{80.69} - 78.1  & noun-adposition & he-htb & 99.71 - \textbf{99.77} - 99.55  \\
subject-verb & hy-armtdp & 81.25 - 80.25 - \textbf{80.25}  & noun-adposition & cu-proiel & 98.06 - 98.4 - \textbf{98.4}  \\
subject-verb & en-ewt & 98.92 - \textbf{98.81} - 94.15  & noun-adposition & sv-lines & 98.6 - 98.11 - \textbf{98.11}  \\
subject-verb & fr-gsd & 96.7 - 94.21 - \textbf{94.21}  & noun-adposition & uk-iu & 99.74 - \textbf{99.8} - 99.54  \\
subject-verb & el-gdt & 77.04 - \textbf{77.93} - 73.56  & noun-adposition & lzh-kyoto & 95.58 - 96.61 - \textbf{96.61}  \\
subject-verb & es-gsd & 79.15 - \textbf{84.14} - 71.52  & noun-adposition & cop-scriptorium & 99.92 - \textbf{99.78} - 99.18  \\
subject-verb & tr-imst & 91.12 - 92.96 - \textbf{92.96}  & noun-adposition & lv-lvtb & 98.56 - 97.78 - \textbf{97.78}  \\
subject-verb & ru-syntagrus & 72.33 - \textbf{80.49} - 72.94  & noun-adposition & et-edt & 98.92 - \textbf{98.77} - 81.84  \\
subject-verb & sl-ssj & 70.95 - \textbf{74.66} - 63.01  & noun-adposition & fro-srcmf & 99.75 - 99.42 - \textbf{99.42}  \\
subject-verb & id-gsd & 99.09 - 99.34 - \textbf{99.34}  & noun-adposition & hy-armtdp & 97.22 - \textbf{96.83} - 85.71  \\
subject-verb & lt-alksnis & 74.44 - \textbf{78.39} - 75.33  & noun-adposition & en-ewt & 99.67 - 99.42 - \textbf{99.42}  \\
subject-verb & ar-nyuad & 91.01 - \textbf{91.32} - 87.82  & noun-adposition & es-gsd & 99.81 - \textbf{100.0} - 98.83  \\
subject-verb & grc-proiel & 69.46 - \textbf{72.23} - 65.71  & noun-adposition & ru-syntagrus & 99.24 - \textbf{99.41} - 99.13  \\
subject-verb & de-hdt & 68.1 - \textbf{76.23} - 61.84  & noun-adposition & id-gsd & 97.67 - \textbf{97.81} - 96.81  \\
numeral-noun & it-vit & 73.17 - 79.32 - \textbf{79.32}  & noun-adposition & ar-nyuad & 99.84 - \textbf{99.87} - 99.48  \\
numeral-noun & no-nynorsk & 88.49 - 88.44 - \textbf{88.44}  & noun-adposition & grc-proiel & 99.03 - 98.92 - \textbf{98.92}  \\
numeral-noun & ro-nonstandard & 87.27 - \textbf{84.83} - 62.07  & noun-adposition & de-hdt & 99.98 - \textbf{99.98} - 99.37  \\
numeral-noun & bg-btb & 92.22 - 88.24 - \textbf{88.24} \\
numeral-noun & cs-pdt & 84.4 - \textbf{88.65} - 69.59 \\
numeral-noun & fi-tdt & 82.35 - \textbf{87.25} - 68.3 \\
numeral-noun & pl-pdb & 97.27 - 97.27 - \textbf{97.27} \\
numeral-noun & la-ittb & 88.0 - \textbf{87.16} - 53.21 \\
numeral-noun & nl-alpino & 95.03 - \textbf{98.7} - 89.61 \\
numeral-noun & mt-mudt & 69.77 - 70.77 - \textbf{70.77} \\
numeral-noun & wo-wtb & 74.63 - \textbf{82.5} - 73.75 \\
numeral-noun & ja-bccwj & 99.05 - 98.71 - \textbf{98.71} \\
numeral-noun & orv-torot & 86.64 - \textbf{79.8} - 72.73 \\
numeral-noun & he-htb & 85.21 - \textbf{80.0} - 64.0 \\
numeral-noun & pt-gsd & 92.18 - \textbf{89.42} - 73.56 \\
numeral-noun & sv-lines & 81.3 - 85.48 - \textbf{85.48} \\
numeral-noun & ga-idt & 73.2 - \textbf{62.86} - 57.14 \\
numeral-noun & sk-snk & 88.01 - \textbf{75.36} - 43.12 \\
numeral-noun & hr-set & 95.39 - \textbf{97.28} - 96.94 \\
numeral-noun & et-edt & 91.63 - \textbf{91.54} - 83.65 \\
numeral-noun & en-ewt & 85.33 - \textbf{89.05} - 82.09 \\
numeral-noun & fr-gsd & 79.7 - \textbf{81.88} - 60.87 \\
numeral-noun & el-gdt & 88.2 - 80.6 - \textbf{80.6} \\
numeral-noun & es-gsd & 87.17 - \textbf{89.43} - 75.61 \\
numeral-noun & ru-syntagrus & 93.48 - \textbf{95.01} - 85.15 \\
numeral-noun & sl-ssj & 84.08 - 78.45 - \textbf{78.45} \\
numeral-noun & id-gsd & 61.04 - \textbf{68.12} - 53.44 \\
numeral-noun & ar-nyuad & 88.96 - \textbf{91.79} - 47.9 \\
numeral-noun & grc-proiel & 68.76 - 62.9 - \textbf{62.9} \\
noun-adposition & no-nynorsk & 99.31 - \textbf{99.26} - 99.14 \\
noun-adposition & gl-ctg & 99.33 - \textbf{99.32} - 99.18 \\
noun-adposition & cs-pdt & 99.98 - \textbf{99.98} - 99.94 \\
  \end{tabular}
    }
    \caption{Accuracy results for all relations across different languages. Baseline is the most frequent order in the training data.}
    \label{tab:wordorderall-2}
    \vspace{-1em}
\end{table*}
\begin{table*}[t]
\small
    \centering
    \resizebox{\textwidth}{!}{
    \begin{tabular}{c|c|c||c|c|c}
    \toprule
    Type & Lang &  Test - Baseline & Type & Lang &  Test - Baseline  \\
    \midrule 

Gender & it-vit & \textbf{71.01} - 67.19 & Gender & hr-set & 71.32 - \textbf{72.5} \\
Person & it-vit & \textbf{70.83} - 62.5 & Person & hr-set & 63.33 - \textbf{76.92} \\
Number & it-vit & 59.56 - \textbf{71.2} & Number & hr-set & 64.25 - \textbf{67.53} \\
Gender & no-nynorsk & \textbf{70.0} - 46.43 & Gender & lv-lvtb & \textbf{74.48} - 72.66 \\
Number & no-nynorsk & 70.0 - \textbf{70.21} & Person & lv-lvtb & \textbf{60.98} - 50.0 \\
Gender & ro-nonstandard & 61.05 - \textbf{63.95} & Number & lv-lvtb & \textbf{72.78} - 68.83 \\
Person & ro-nonstandard & 55.22 - \textbf{63.64} & Gender & hsb-ufal & 60.87 - \textbf{85.71} \\
Number & ro-nonstandard & \textbf{62.86} - 62.63 & Number & hsb-ufal & 46.72 - \textbf{69.23} \\
Gender & bg-btb & \textbf{66.0} - 63.83 & Gender & ru-syntagrus & 64.96 - \textbf{69.68} \\
Person & bg-btb & \textbf{64.0} - 62.5 & Person & ru-syntagrus & \textbf{64.0} - 62.5 \\
Number & bg-btb & \textbf{73.17} - 63.93 & Number & ru-syntagrus & \textbf{60.24} - 59.09 \\
Gender & cs-pdt & \textbf{75.09} - 56.44 & Gender & el-gdt & \textbf{73.58} - 63.83 \\
Person & cs-pdt & 57.78 - \textbf{59.09} & Person & el-gdt & 65.0 - \textbf{66.67} \\
Number & cs-pdt & \textbf{63.35} - 47.66 & Number & el-gdt & \textbf{76.54} - 62.73 \\
Gender & pl-pdb & \textbf{71.11} - 64.53 & Gender & hi-hdtb & \textbf{69.11} - 58.59 \\
Person & pl-pdb & \textbf{60.71} - 55.56 & Number & hi-hdtb & \textbf{71.77} - 41.61 \\
Number & pl-pdb & \textbf{66.06} - 63.68 & Gender & es-gsd & \textbf{84.31} - 71.83 \\
Gender & la-ittb & \textbf{77.78} - 73.53 & Person & es-gsd & \textbf{91.67} - 59.09 \\
Person & la-ittb & 19.05 - \textbf{19.05} & Number & es-gsd & \textbf{88.89} - 64.39 \\
Number & la-ittb & \textbf{65.14} - 57.89 & Gender & ta-ttb & \textbf{100.0} - 68.18 \\
Gender & nl-alpino & 56.25 - \textbf{66.67} & Number & ta-ttb & \textbf{77.78} - 52.27 \\
Number & nl-alpino & \textbf{60.94} - 54.84 & Person & ug-udt & 37.93 - \textbf{52.63} \\
Gender & orv-torot & 64.52 - \textbf{65.54} & Number & ug-udt & 47.73 - \textbf{76.67} \\
Person & orv-torot & \textbf{66.67} - 60.0 & Person & fi-tdt & \textbf{58.06} - 38.71 \\
Number & orv-torot & \textbf{64.04} - 62.12 & Number & fi-tdt & \textbf{60.0} - 50.23 \\
Gender & he-htb & \textbf{78.16} - 74.7 & Person & wo-wtb & 52.17 - \textbf{55.0} \\
Person & he-htb & \textbf{78.95} - 73.68 & Number & wo-wtb & \textbf{57.14} - 48.57 \\
Number & he-htb & 58.14 - \textbf{58.54} & Person & hu-szeged & 39.39 - \textbf{44.44} \\
Gender & cu-proiel & 58.26 - \textbf{61.0} & Number & hu-szeged & 38.34 - \textbf{39.63} \\
Person & cu-proiel & 61.54 - \textbf{66.67} & Person & et-edt & \textbf{68.75} - 61.29 \\
Number & cu-proiel & 60.4 - \textbf{67.21} & Number & et-edt & 61.21 - \textbf{64.84} \\
Gender & mr-ufal & 53.57 - \textbf{60.87} & Person & hy-armtdp & \textbf{57.14} - 44.44 \\
Person & mr-ufal & 28.57 - \textbf{72.73} & Number & hy-armtdp & 58.49 - \textbf{59.18} \\
Number & mr-ufal & \textbf{66.67} - 39.39 & Person & en-ewt & \textbf{100.0} - 81.25 \\
Gender & be-hse & \textbf{61.29} - 59.57 & Number & en-ewt & \textbf{69.0} - 35.71 \\
Number & be-hse & \textbf{65.82} - 64.62 & Person & tr-imst & 32.69 - \textbf{35.91} \\
Gender & sv-lines & \textbf{65.52} - 53.85 & Number & tr-imst & \textbf{84.62} - 46.96 \\
Number & sv-lines & 60.0 - \textbf{64.29} & Number & kmr-mg & 55.56 - \textbf{78.26} \\
Gender & uk-iu & 68.92 - \textbf{70.08} & Number & af-afribooms & \textbf{68.75} - 60.0 \\
Person & uk-iu & \textbf{72.73} - 70.0 & Number & fr-gsd & \textbf{75.0} - 62.37 \\
Number & uk-iu & \textbf{65.78} - 64.67\\
Gender & ga-idt & \textbf{73.77} - 64.0\\
Person & ga-idt & 42.86 - \textbf{62.5}\\
Number & ga-idt & 43.16 - \textbf{46.75}\\
Gender & sk-snk & \textbf{71.9} - 69.16\\
Person & sk-snk & \textbf{88.89} - 77.78\\
Number & sk-snk & \textbf{63.36} - 55.83\\
Gender & got-proiel & \textbf{62.02} - 55.86\\
Person & got-proiel & \textbf{62.16} - 57.14\\
Number & got-proiel & \textbf{67.51} - 64.0\\

    \end{tabular}
    }
    \caption{Accuracy results for all relations across different languages.  Baseline is \citet{chaudhary-etal-2020-automatic}}
    \label{tab:agreementall-1}
    \vspace{-1em}
\end{table*}
\begin{table*}[t]
\small
    \centering
    \resizebox{\textwidth}{!}{
    \begin{tabular}{c|c|c||c|c|c}
    \toprule
    Type & Lang &  Train - Test - Baseline & Type & Lang &  Train  - Test - Baseline  \\
    \midrule 
PRON & no-nynorsk & 98.55 - \textbf{99.55} - 78.28  & VERB & ug-udt & 76.0 - \textbf{75.64} - 71.37  \\
PRON & ug-udt & 92.22 - \textbf{94.87} - 73.68  & VERB & got-proiel & 85.51 - \textbf{86.15} - 81.15  \\
PRON & ro-nonstandard & 89.77 - \textbf{91.2} - 38.33  & VERB & lv-lvtb & 96.43 - \textbf{95.61} - 75.58  \\
PRON & sk-snk & 83.19 - \textbf{83.9} - 34.75  & VERB & tr-imst & 67.53 - \textbf{66.58} - 46.13  \\
PRON & hu-szeged & 73.94 - \textbf{79.15} - 59.46  & VERB & et-edt & 86.95 - \textbf{86.08} - 82.91  \\
PRON & got-proiel & 87.97 - \textbf{91.05} - 36.79  & VERB & hy-armtdp & 86.63 - \textbf{94.34} - 39.62  \\
PRON & hr-set & 88.6 - \textbf{89.54} - 68.79  & VERB & ur-udtb & 96.01 - 98.95 - \textbf{98.95}  \\
PRON & lv-lvtb & 90.64 - \textbf{90.85} - 54.03  & VERB & lt-alksnis & 94.86 - \textbf{95.0} - 52.5  \\
PRON & en-ewt & 97.74 - \textbf{96.76} - 81.48  & ADP & ro-nonstandard & 98.5 - 98.85 - \textbf{98.85}  \\
PRON & el-gdt & 93.5 - \textbf{93.35} - 36.8  & ADP & sk-snk & 41.74 - \textbf{44.46} - 40.74  \\
PRON & tr-imst & 71.0 - \textbf{73.33} - 42.5  & ADP & hr-set & 45.85 - \textbf{48.42} - 37.96  \\
PRON & sme-giella & 85.31 - \textbf{76.82} - 47.05  & ADP & hi-hdtb & 85.57 - \textbf{86.99} - 52.34  \\
PRON & es-gsd & 95.89 - \textbf{96.14} - 53.71  & ADP & ur-udtb & 82.06 - \textbf{96.59} - 63.54  \\
PRON & da-ddt & 84.38 - \textbf{82.53} - 54.7  & ADP & uk-iu & 45.85 - \textbf{43.39} - 32.85  \\
PRON & et-edt & 79.75 - \textbf{81.58} - 45.26  & ADJ & ro-nonstandard & 98.14 - \textbf{96.9} - 96.42  \\
PRON & af-afribooms & 58.86 - \textbf{53.07} - 31.2  & ADJ & ga-idt & 95.47 - \textbf{93.25} - 90.18  \\
PRON & hy-armtdp & 78.1 - \textbf{79.05} - 63.81  & ADJ & sk-snk & 99.03 - \textbf{98.71} - 35.01  \\
PRON & mr-ufal & 71.58 - 78.95 - \textbf{78.95}  & ADJ & hu-szeged & 98.73 - \textbf{98.25} - 92.58  \\
PRON & be-hse & 81.3 - \textbf{76.12} - 65.67  & ADJ & got-proiel & 88.48 - \textbf{92.33} - 38.36  \\
PRON & ur-udtb & 87.78 - \textbf{90.53} - 54.73  & ADJ & hr-set & 97.75 - \textbf{98.3} - 37.5  \\
PRON & lt-alksnis & 82.8 - \textbf{80.28} - 30.28  & ADJ & lv-lvtb & 93.85 - \textbf{94.37} - 39.59  \\
PRON & bg-btb & 95.73 - \textbf{95.78} - 46.78  & ADJ & et-edt & 94.13 - \textbf{95.16} - 41.07  \\
PRON & sv-lines & 99.32 - \textbf{99.41} - 58.02  & ADJ & el-gdt & 85.61 - \textbf{89.49} - 48.6  \\
PRON & uk-iu & 88.52 - \textbf{90.98} - 48.82  & ADJ & hi-hdtb & 84.36 - \textbf{84.34} - 70.48  \\
NOUN & ug-udt & 78.31 - \textbf{77.13} - 63.46  & ADJ & tr-imst & 56.45 - \textbf{60.22} - 51.88  \\
NOUN & ro-nonstandard & 96.68 - \textbf{97.53} - 87.8  & ADJ & sme-giella & 86.09 - 90.55 - \textbf{90.55}  \\
NOUN & kmr-mg & 53.09 - 47.21 - \textbf{47.21}  & ADJ & ar-nyuad & 94.15 - \textbf{96.94} - 62.54  \\
NOUN & ga-idt & 93.26 - \textbf{95.62} - 80.28  & ADJ & be-hse & 89.59 - \textbf{95.06} - 43.83  \\
NOUN & sk-snk & 91.27 - \textbf{92.27} - 20.61  & ADJ & ur-udtb & 99.04 - \textbf{98.81} - 62.02  \\
NOUN & hu-szeged & 72.25 - \textbf{72.1} - 47.6  & ADJ & lt-alksnis & 96.89 - \textbf{96.72} - 25.5  \\
NOUN & got-proiel & 84.89 - \textbf{87.12} - 27.72  & ADJ & uk-iu & 97.38 - \textbf{98.15} - 46.39  \\
NOUN & hr-set & 88.35 - \textbf{92.21} - 34.41  & DET & ro-nonstandard & 97.01 - \textbf{95.87} - 75.58  \\
NOUN & lzh-kyoto & 89.86 - \textbf{93.72} - 76.61  & DET & sk-snk & 95.7 - \textbf{93.24} - 43.74  \\
NOUN & lv-lvtb & 81.94 - \textbf{83.87} - 31.62  & DET & got-proiel & 94.78 - \textbf{96.25} - 32.29  \\
NOUN & kk-ktb & 49.24 - 53.32 - \textbf{53.32}  & DET & hr-set & 94.87 - \textbf{95.64} - 42.81  \\
NOUN & et-edt & 62.6 - \textbf{66.51} - 27.65  & DET & lv-lvtb & 96.59 - \textbf{97.12} - 30.15  \\
NOUN & el-gdt & 91.02 - \textbf{94.69} - 49.72  & DET & et-edt & 96.73 - \textbf{96.19} - 34.25  \\
NOUN & hi-hdtb & 96.1 - \textbf{97.35} - 54.72  & DET & el-gdt & 91.42 - \textbf{93.64} - 47.48  \\
NOUN & tr-imst & 59.03 - \textbf{64.52} - 54.65  & DET & hi-hdtb & 88.89 - \textbf{92.87} - 76.1  \\
NOUN & ta-ttb & 77.24 - \textbf{76.49} - 68.02  & DET & ur-udtb & 95.79 - \textbf{95.91} - 64.33  \\
NOUN & sme-giella & 76.51 - \textbf{78.78} - 30.32  & DET & lt-alksnis & 79.46 - \textbf{83.65} - 39.92  \\
NOUN & ar-nyuad & 87.66 - \textbf{94.66} - 67.49  & DET & uk-iu & 94.31 - \textbf{94.86} - 27.93  \\
NOUN & hsb-ufal & 24.07 - 19.53 - \textbf{19.53}  & PROPN & ro-nonstandard & 97.35 - \textbf{96.77} - 92.98  \\
NOUN & hy-armtdp & 78.17 - \textbf{80.2} - 46.08  & PROPN & ga-idt & 79.87 - \textbf{85.78} - 73.28  \\
NOUN & mr-ufal & 81.38 - \textbf{75.0} - 42.65  & PROPN & sk-snk & 90.24 - \textbf{88.9} - 46.39  \\
NOUN & be-hse & 69.27 - \textbf{75.95} - 46.1  & PROPN & hu-szeged & 91.47 - 89.36 - \textbf{89.36}  \\
NOUN & ur-udtb & 92.1 - \textbf{96.81} - 51.25  & PROPN & got-proiel & 85.91 - \textbf{86.89} - 50.91  \\
NOUN & lt-alksnis & 85.08 - \textbf{82.93} - 39.07  & PROPN & hr-set & 92.42 - \textbf{94.67} - 48.27  \\
NOUN & sv-lines & 99.6 - \textbf{99.86} - 97.47  & PROPN & lv-lvtb & 88.64 - \textbf{90.13} - 39.91  \\
NOUN & uk-iu & 94.1 - \textbf{94.73} - 43.79  & PROPN & el-gdt & 91.44 - \textbf{90.32} - 32.58  \\

  \end{tabular}
    }
    \caption{Accuracy results for all relations across different languages.  Baseline is the most frequent case value in the training data.}
    \label{tab:assignmentall-1}
    \vspace{-1em}
\end{table*}

\begin{table*}[t]
\small
    \centering
    \resizebox{\textwidth}{!}{
    \begin{tabular}{c|c|c||c|c|c}
    \toprule
    Type & Lang &  Train - Test - Baseline & Type & Lang &  Train - Test - Baseline  \\
    \midrule 

VERB & ug-udt & 76.0 - \textbf{75.64} - 71.37  & PROPN & hi-hdtb & 94.91 - \textbf{96.49} - 48.51  \\
VERB & got-proiel & 85.51 - \textbf{86.15} - 81.15  & PROPN & tr-imst & 73.55 - \textbf{71.73} - 68.0  \\
VERB & lv-lvtb & 96.43 - \textbf{95.61} - 75.58  & PROPN & ta-ttb & 97.99 - \textbf{94.84} - 93.55  \\
VERB & tr-imst & 67.53 - \textbf{66.58} - 46.13  & PROPN & sme-giella & 84.23 - \textbf{82.9} - 35.81  \\
VERB & et-edt & 86.95 - \textbf{86.08} - 82.91  & PROPN & ar-nyuad & 78.68 - \textbf{84.27} - 59.85  \\
VERB & hy-armtdp & 86.63 - \textbf{94.34} - 39.62  & PROPN & et-edt & 75.05 - \textbf{83.18} - 51.24  \\
VERB & ur-udtb & 96.01 - 98.95 - \textbf{98.95}  & PROPN & hy-armtdp & 82.28 - \textbf{89.13} - 54.89  \\
VERB & lt-alksnis & 94.86 - \textbf{95.0} - 52.5  & PROPN & be-hse & 86.43 - 72.68 - \textbf{72.68}  \\
ADP & ro-nonstandard & 98.5 - 98.85 - \textbf{98.85}  & PROPN & ur-udtb & 92.7 - \textbf{97.65} - 59.77  \\
ADP & sk-snk & 41.74 - \textbf{44.46} - 40.74  & PROPN & sv-lines & 97.21 - \textbf{96.6} - 91.23  \\
ADP & hr-set & 45.85 - \textbf{48.42} - 37.96  & PROPN & uk-iu & 93.76 - \textbf{95.14} - 36.14  \\
ADP & hi-hdtb & 85.57 - \textbf{86.99} - 52.34  & NUM & sk-snk & 81.47 - \textbf{77.38} - 39.29  \\
ADP & ur-udtb & 82.06 - \textbf{96.59} - 63.54  & NUM & got-proiel & 44.0 - \textbf{45.83} - 33.33  \\
ADP & uk-iu & 45.85 - \textbf{43.39} - 32.85  & NUM & hr-set & 90.27 - \textbf{94.26} - 41.8  \\
ADJ & ro-nonstandard & 98.14 - \textbf{96.9} - 96.42  & NUM & lv-lvtb & 88.07 - \textbf{85.44} - 38.61  \\
ADJ & ga-idt & 95.47 - \textbf{93.25} - 90.18  & NUM & el-gdt & 75.75 - \textbf{73.17} - 58.54  \\
ADJ & sk-snk & 99.03 - \textbf{98.71} - 35.01  & NUM & tr-imst & 76.55 - \textbf{82.22} - 77.78  \\
ADJ & hu-szeged & 98.73 - \textbf{98.25} - 92.58  & NUM & sme-giella & 47.8 - 41.84 - \textbf{41.84}  \\
ADJ & got-proiel & 88.48 - \textbf{92.33} - 38.36  & NUM & et-edt & 88.9 - \textbf{93.51} - 70.3  \\
ADJ & hr-set & 97.75 - \textbf{98.3} - 37.5  & NUM & uk-iu & 90.46 - \textbf{92.48} - 52.29  \\
ADJ & lv-lvtb & 93.85 - \textbf{94.37} - 39.59  & ADV & fa-seraji & 85.35 - 81.36 - \textbf{81.36}  \\
ADJ & et-edt & 94.13 - \textbf{95.16} - 41.07 \\
ADJ & el-gdt & 85.61 - \textbf{89.49} - 48.6 \\
ADJ & hi-hdtb & 84.36 - \textbf{84.34} - 70.48 \\
ADJ & tr-imst & 56.45 - \textbf{60.22} - 51.88 \\
ADJ & sme-giella & 86.09 - 90.55 - \textbf{90.55} \\
ADJ & ar-nyuad & 94.15 - \textbf{96.94} - 62.54 \\
ADJ & be-hse & 89.59 - \textbf{95.06} - 43.83 \\
ADJ & ur-udtb & 99.04 - \textbf{98.81} - 62.02 \\
ADJ & lt-alksnis & 96.89 - \textbf{96.72} - 25.5 \\
ADJ & uk-iu & 97.38 - \textbf{98.15} - 46.39 \\
DET & ro-nonstandard & 97.01 - \textbf{95.87} - 75.58 \\
DET & sk-snk & 95.7 - \textbf{93.24} - 43.74 \\
DET & got-proiel & 94.78 - \textbf{96.25} - 32.29 \\
DET & hr-set & 94.87 - \textbf{95.64} - 42.81 \\
DET & lv-lvtb & 96.59 - \textbf{97.12} - 30.15 \\
DET & et-edt & 96.73 - \textbf{96.19} - 34.25 \\
DET & el-gdt & 91.42 - \textbf{93.64} - 47.48 \\
DET & hi-hdtb & 88.89 - \textbf{92.87} - 76.1 \\
DET & ur-udtb & 95.79 - \textbf{95.91} - 64.33 \\
DET & lt-alksnis & 79.46 - \textbf{83.65} - 39.92 \\
DET & uk-iu & 94.31 - \textbf{94.86} - 27.93 \\
PROPN & ro-nonstandard & 97.35 - \textbf{96.77} - 92.98 \\
PROPN & ga-idt & 79.87 - \textbf{85.78} - 73.28 \\
PROPN & sk-snk & 90.24 - \textbf{88.9} - 46.39 \\
PROPN & hu-szeged & 91.47 - 89.36 - \textbf{89.36} \\
PROPN & got-proiel & 85.91 - \textbf{86.89} - 50.91 \\
PROPN & hr-set & 92.42 - \textbf{94.67} - 48.27 \\
PROPN & lv-lvtb & 88.64 - \textbf{90.13} - 39.91 \\
PROPN & el-gdt & 91.44 - \textbf{90.32} - 32.58 \\

  \end{tabular}
    }
    \caption{Accuracy results for all relations across different languages.  Baseline is the most frequent value training data.}
    \label{tab:assignmentall-2}
    \vspace{-1em}
\end{table*}
\end{document}